\documentclass[14pt]{article}
\usepackage{arxiv}

\usepackage[utf8]{inputenc} 
\usepackage[T1]{fontenc}    
\usepackage{hyperref}       
\usepackage{url}            
\usepackage{booktabs}       
\usepackage{amsfonts}       
\usepackage{nicefrac}       
\usepackage{microtype}      
\usepackage{lipsum}		
\usepackage{graphicx}
\usepackage{doi}

\usepackage{amsmath}
\usepackage{times}
\usepackage{color}
\usepackage{lscape}
\usepackage[authoryear]{natbib}

\usepackage{amssymb}
\usepackage{subcaption}
\usepackage{float}

\usepackage{multirow}
\usepackage{xcolor}

\usepackage{tabularx}
\usepackage{adjustbox}
\usepackage{mwe}

\usepackage{xr}
\makeatletter
\newcommand*{\addFileDependency}[1]{
  \typeout{(#1)}
  \@addtofilelist{#1}
  \IfFileExists{#1}{}{\typeout{No file #1.}}
}
\makeatother

\usepackage{subfiles}

\usepackage{authblk}

\title{Understanding the computational demands underlying visual reasoning}
\date{}

\author[1,2]{\bf \large Mohit Vaishnav}
\author[2]{\bf \large Remi Cadene}
\author[3]{\bf \large Andrea Alamia}
\author[2]{\bf \large Drew Linsley}
\author[1,3]{\bf \large Rufin VanRullen}
\author[1,2]{\bf \large Thomas Serre}
\affil[1]{Artificial and Natural Intelligence Toulouse Institute, Université de Toulouse, France}
\affil[2]{Carney Institute for Brain Science, Dpt. of Cognitive Linguistic \& Psychological Sciences,
Brown University, Providence, RI 02912}
\affil[3]{Centre de Recherche Cerveau \& Cognition (CerCo), CNRS, Universite de Toulouse, 31052 Toulouse, France}

\begin{document}
\maketitle



\begin{center} {\bf Abstract } \end{center}
Visual understanding requires comprehending complex visual relations between objects within a scene. Here,  we seek to characterize the computational demands for abstract visual reasoning. We do this by systematically assessing the  ability of modern deep convolutional neural networks (CNNs) to learn to solve the ``Synthetic Visual Reasoning Test" (SVRT) challenge, a collection of twenty-three visual reasoning problems. 
Our analysis reveals a novel taxonomy of visual reasoning tasks, which can be primarily explained by both the type of relations (same-different vs. spatial-relation judgments) and the number of relations used to compose the underlying rules. 
Prior cognitive neuroscience work suggests that attention plays a key role in \textcolor{black}{humans'} visual reasoning ability. To test this \textcolor{black}{hypothesis}, we extended the CNNs with spatial and feature-based attention mechanisms. In a second series of experiments, we evaluated the ability of these attention networks to learn to solve the SVRT challenge and found the resulting architectures to be much more efficient at solving the hardest of these visual reasoning tasks. Most importantly, the corresponding improvements on individual tasks partially explained our novel taxonomy. Overall, this work provides an granular computational account of visual reasoning and yields testable \textcolor{black}{neuroscience} predictions regarding the differential need for feature-based vs. spatial attention \textcolor{black}{depending on the type of visual reasoning problem}. 

\section*{Introduction}

Humans can effortlessly reason about the visual world and provide rich and detailed descriptions of briefly presented real-life photographs \citep{Fei-Fei2007-zb}, vastly outperforming the best current computer vision systems \citep{Geman2015-jm, Kreiman2020-zd}. 
For the most part, studies of visual reasoning in humans have sought to characterize the neural computations underlying the judgment of individual relations between objects, such as their spatial relations (e.g., \citet{logan1994ability}) or whether they are the same or different (up to a transformation, e.g., \citet{shepard1971mental}). It has also been shown that different visual reasoning problems have different attentional and working memory demands \citep{logan1994spatial,Moore1994,Rosielle2002,Holcombe2011,vanderham2012,kroger2002recruitment,Golde2010,clevenger2014working,brady2015contextual}. However, there is still little known about the neural computations that are engaged by different types of visual reasoning (see~\citet{ricci37same} for a recent review). 

\begin{figure}[t]
\centering
  \includegraphics[width=.8\linewidth]{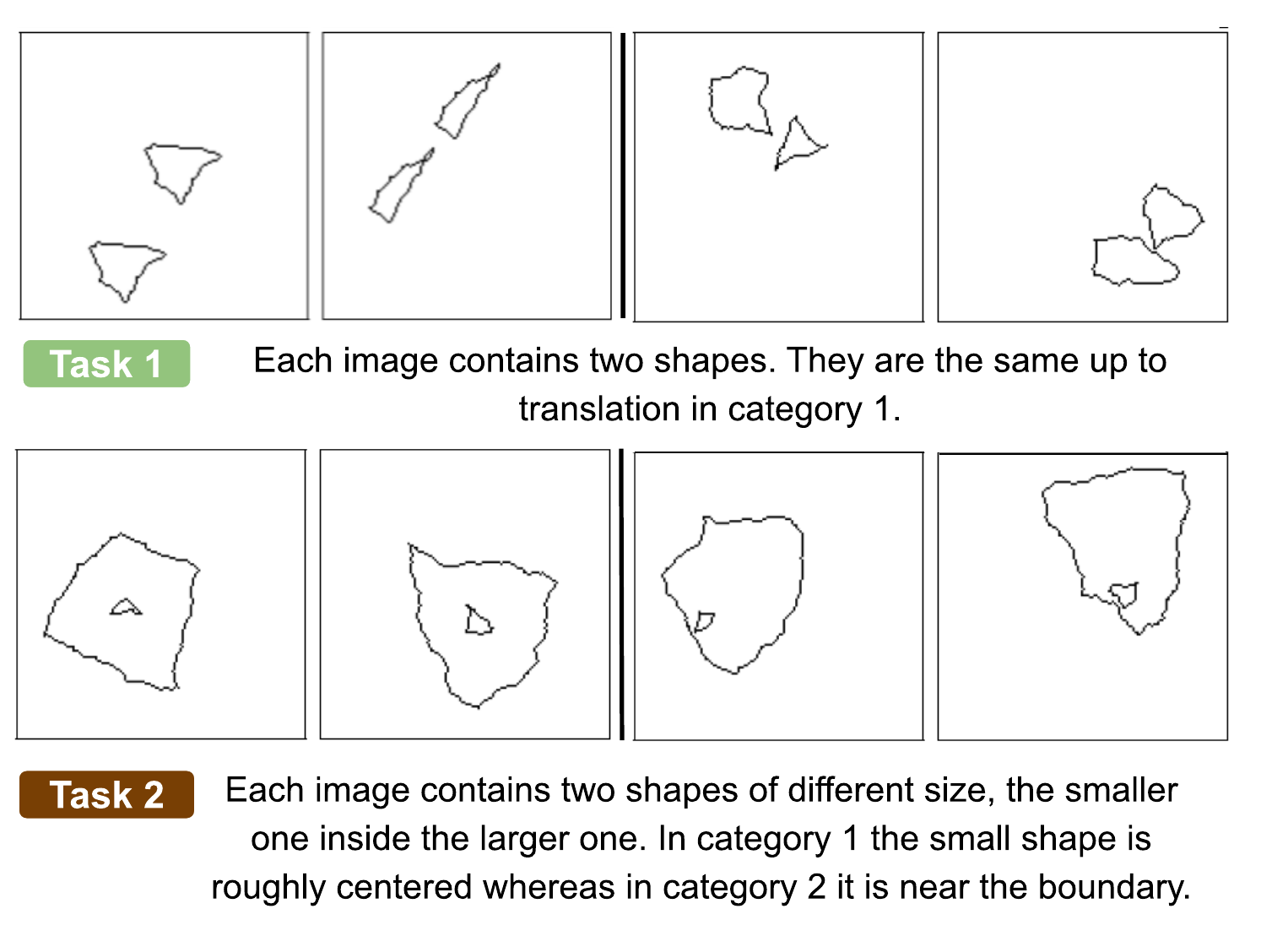}
 \caption{Two SVRT sample tasks from a set of twenty three in total. For each task, the leftmost and rightmost two examples illustrate the two categories to be classified. Representative samples for the complete set of twenty-three tasks can be found in Figure~\ref{fig:exampleSD} and~\ref{fig:exampleSR}.} \label{fig:example}
\end{figure}

One benchmark that has been designed to probe abstract visual relational capabilities in humans and machines is the \textit{Synthetic Visual Reasoning Test} (SVRT) ~\citep{fleuret2011comparing}. The dataset consists of twenty-three hand-designed binary classification problems that test abstract relationships between objects posed on images of closed-contour shapes. Observers are never explicitly given the underlying rule for solving any given problem. Instead, they learn it while classifying positive and negative examples and receiving task feedback. Examples from two representative tasks are depicted in Figure~\ref{fig:example}: observers must learn to recognize whether two shapes are the same or different (Task \textit{1}), or whether or not the smaller of the two shapes is near the boundary (Task \textit{2}). Additional abstract relationships tested in the challenge include ``inside", ``in between”, ``forming a square”, ``aligned in a row" or ``finding symmetry" (see Figures \ref{fig:exampleSD} and \ref{fig:exampleSR} for examples).

Most SVRT tasks are rapidly learned by human observers within twenty or fewer training examples~\citep{fleuret2011comparing} (see Table~\ref{table:humans}; reproduced from the original study). On the other hand, modern deep neural network models require several orders of magnitude more training samples for some of the more challenging tasks~\citep{ellis2015unsupervised,stabinger201625,kim2018not,MESSINA2021,stabinger2021evaluating,Stabinger2016,Puebla2021.04.06.438551} (see~\citet{ricci37same} for review; see also \citet{Funke2021} for an alternative perspective). 

It is now clear that some SVRT tasks are more difficult to learn than others. For instance, tasks that involve spatial-relation (SR) judgments can be learned much more easily by deep convolutional neural networks (CNNs) than tasks that involve same-different (SD) judgments~\citep{Stabinger2016, kim2018not,yihe2019program}. In contrast, a very recent study \citep{Puebla2021.04.06.438551} demonstrated that even when CNNs learn to detect whether objects are the same or different, they fail to generalize over small changes in appearance, meaning that they have only partially learned this abstract rule. The implication of the relative difficulty of learning SR versus SD tasks is that CNNs appear to need additional computations to solve SD tasks beyond standard filtering, non-linear rectification, and pooling. Indeed, recent human electrophysiology work~\citep{AlamiaENEURO.0267-20.2020} has shown that SD tasks recruit cortical mechanisms associated with attention and working memory processes to a greater extent than SR tasks. \textcolor{black}{Others have argued that SD tasks are central to human intelligence \citep{firestone2020performance, FORBUS202163, GENTNER202184}.} Beyond this basic dichotomy of SR and SD tasks, little is known about the neural computations necessary to learn to solve SVRT tasks as efficiently as human observers.

Here, we investigate the neural computations required for visual reasoning in two sets of experiments. In our first set of experiments, we extend prior studies on the learnability of individual SVRT tasks by feedforward neural networks using a popular class of deep neural networks known as deep residual networks (``ResNets'') \citep{he2016deep}. We systematically analyze the ability of ResNets to learn all twenty-three SVRT tasks as a function of their expressivness, parameterized by processing depth (number of layers), and their efficiency in learning a particular task. Through these experiments, we found that most of the performance variance in the space of SVRT tasks could be accounted for by two principal components, which reflected both the type of task (same-different vs. spatial-relation judgments) and the number of relations used to compose the underlying rules. 

Consistent with the speculated role of attention in solving the binding problem when reasoning about objects \citep{egly1994covert,roelfsema1998object}, prior work by \citet{kim2018not} has shown that combining CNNs with an oracle model of attention and feature binding (i.e., preprocessing images so that they are explicitly and readily organized into discrete object channels) renders SD tasks as easy to learn by CNNs as SR tasks. Here, we build on this work and introduce CNN extensions that incorporate spatial or feature-based attention.  In a second set of experiments, we show that these attention networks learn difficult SVRT tasks with fewer training examples than their \textcolor{black}{non}-attentive (CNN) counterparts, but that the different forms of attention help on different tasks.

This second set of experiments raises the question: how do attention mechanisms help with learning different visual reasoning problems? There are at least two possible computational benefits: attention could either improve model performance by simply increasing its capacity or attention could help models more efficiently learn the abstract rules governing object relationships. To adjudicate between these two possibilities, we measured the sample efficiency of ResNets pre-trained on SVRT images, so that they only had to learn the abstract rules for each SVRT task. We found that attention ResNets and ResNets pre-trained on the SVRT were similarly sample-efficient in learning new SVRT tasks, indicating that attention helps discover abstract rules instead of merely increasing model capacity.

\section*{Experiment 1: ResNets}
\subsection*{Systematic analysis of SVRT tasks' learnability}

All experiments were carried out with the \textit{Synthetic Visual Reasoning Test} (SVRT) dataset using code provided by the authors to generate images with dimension \textit{128 $\times$ 128} pixels (see \citet{fleuret2011comparing} for details). \textcolor{black}{ All images were normalized and resized to 256$\times$256 pixels for training and testing models. No image augmentations were used during training.}
In our first experiment, we wanted to measure how easy or difficult each task is for ResNets to learn. We did this by recording the SVRT performance of multiple ResNets, each with different numbers of layers and trained with different numbers of examples. By varying model complexity and the number of samples provided to a model to learn any given task, we obtained complementary measures of the learnability of every SVRT task for ResNet architectures. In total, we trained 18-, 50-, and 152-layer ResNets separately on each of the SVRT's twenty-three task. Each of these models was trained with .5k, 1k, 5k, 10k, 15k, and 120k class-balanced samples. We also generated two unique sets of 40k positive and negative samples for each task: one was used as a validation set to select a stopping criterion for training the networks (if validation accuracy reaches 100\%) and one was used as a test set to report model accuracy. In addition, we used three independent random initializations of the training weights for each configuration of architecture/task and selected the best model using the validation set. Models were trained for \textit{100} epochs using the $Adam$ optimizer \citep{kingma2014adam}  with a training schedule (we used an initial learning rate of 1$e$-3 and changing it to 1$e$-4  from the $70^{th}$ epoch \textcolor{black}{onward}). As a control, because these tasks are quite different from each other, we also tested two additional initial learning rates (\textit{1e-4, 1e-5}).

\begin{figure*}[t!]
\centering
  \includegraphics[width=.55\linewidth]{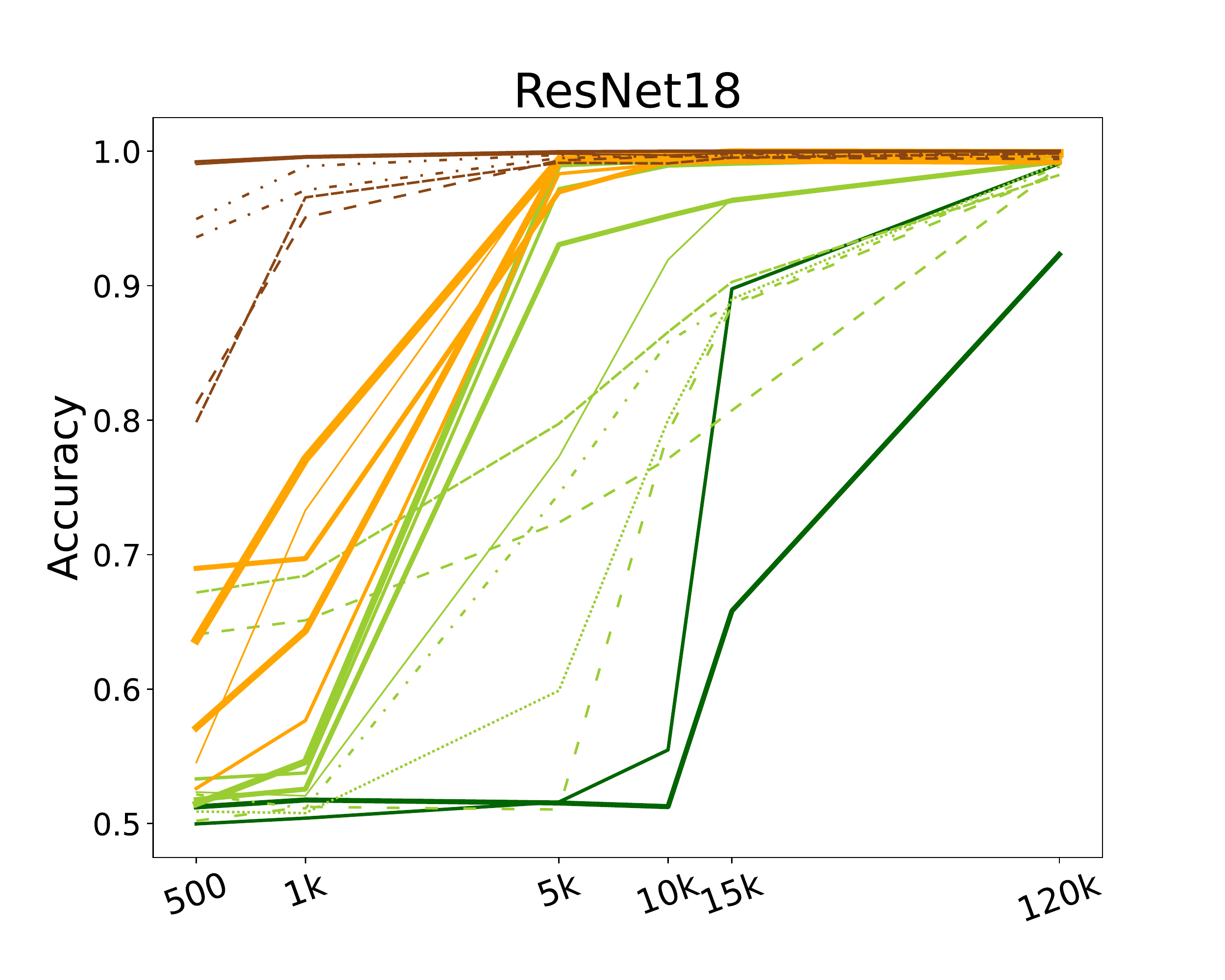} \includegraphics[width=.55\linewidth]{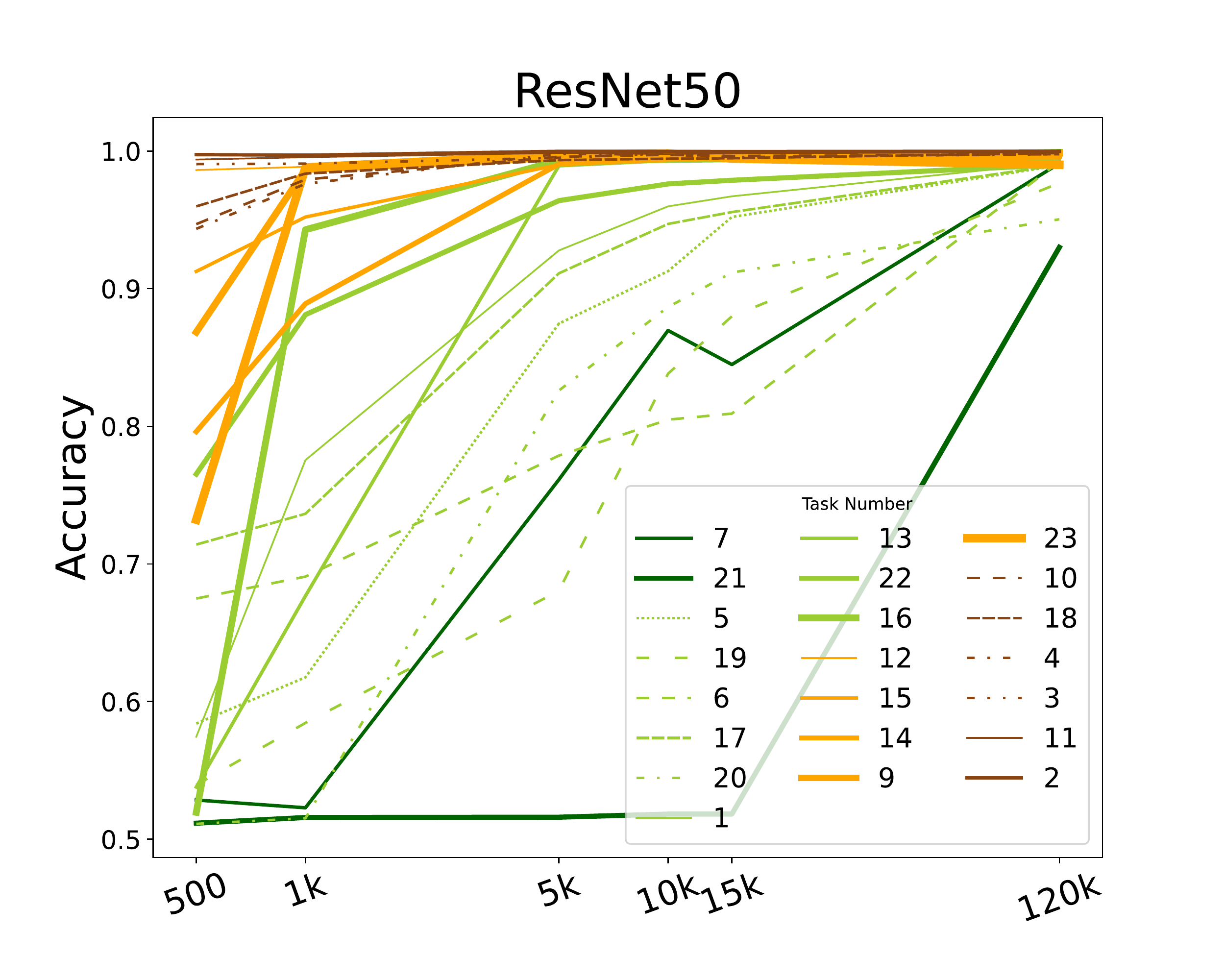}
  \includegraphics[width=.55\linewidth]{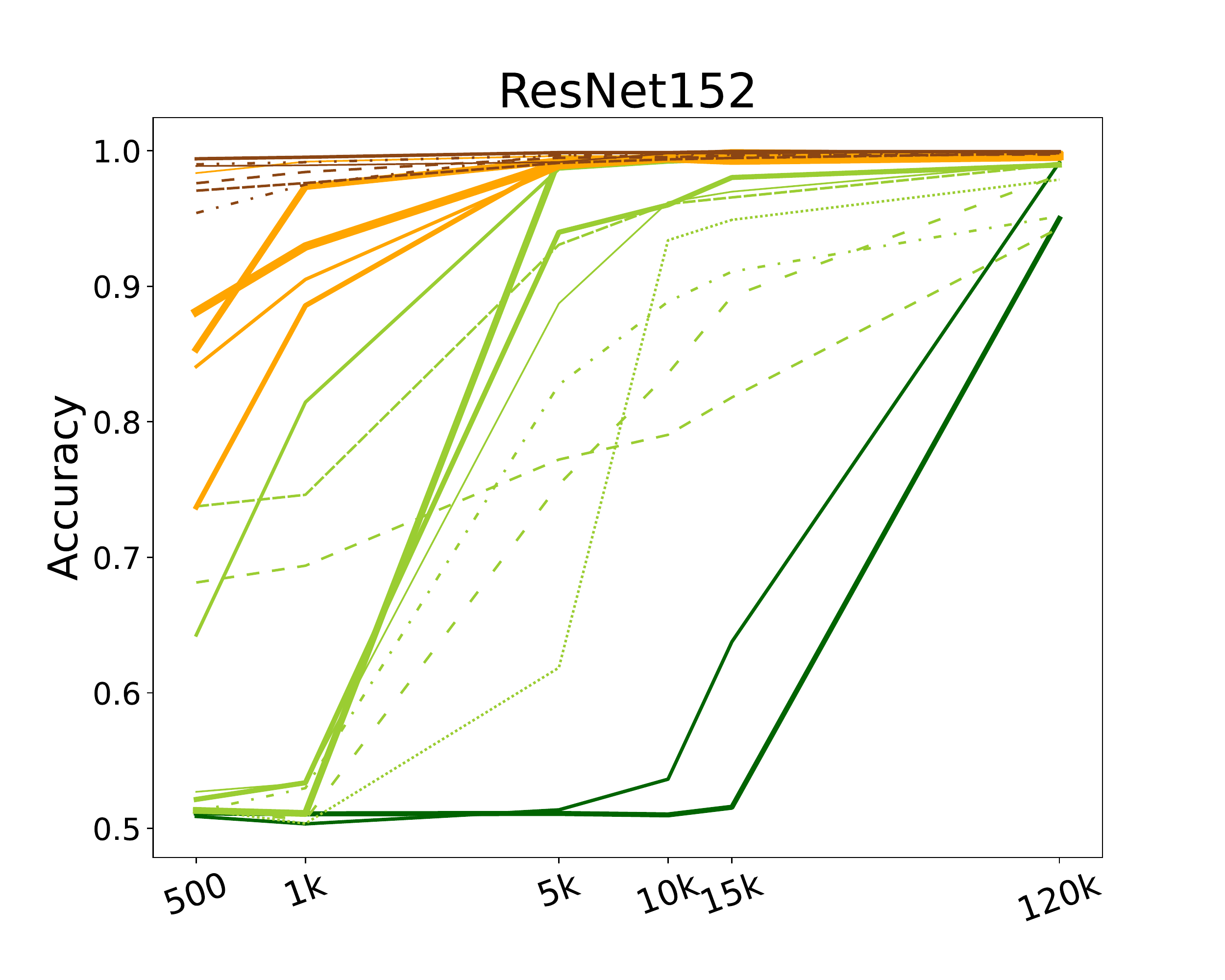}
 \caption{Test accuracy for each of the twenty-three SVRT tasks as a function of the number of training samples for ResNets with depth 18, 50 and 152, resp. The color scheme reflects the identified taxonomy of SVRT tasks (see Figure~\ref{fig:clustering} and text for details).}\label{fig:overall}
\end{figure*}

Consistent with prior work \citep{kim2018not,Stabinger2016,yihe2019program}, we found that some SVRT tasks are much easier to learn than others for ResNets (Figure~\ref{fig:overall}). For instance, a ResNet50 needs only \textit{500} examples to perform well on tasks \textit{2, 3, 4, 8, 10, 11, 18} but the same network needs \textit{120k} samples to perform well on task \textit{21} (see Figures~\ref{fig:exampleSD} and \ref{fig:exampleSR} for examples of these tasks). Similarly, with \textit{500} training examples, task \textit{2, 3, 4 \& 11} can be learned well with only 18 layers while task \textit{9, 12, 15 \& 23} require as many as 152 layers. A key assumption of our work is that these differences in training set sizes and depth requirements between different SVRT tasks reflect different computational strategies that need to be discovered by the neural networks during training for different tasks. Our next goal is to charactarize what these computational strategies are.

\clearpage

\subsection*{An SVRT taxonomy}
\begin{figure}[t!]
\centering
  \includegraphics[width=.95\linewidth]{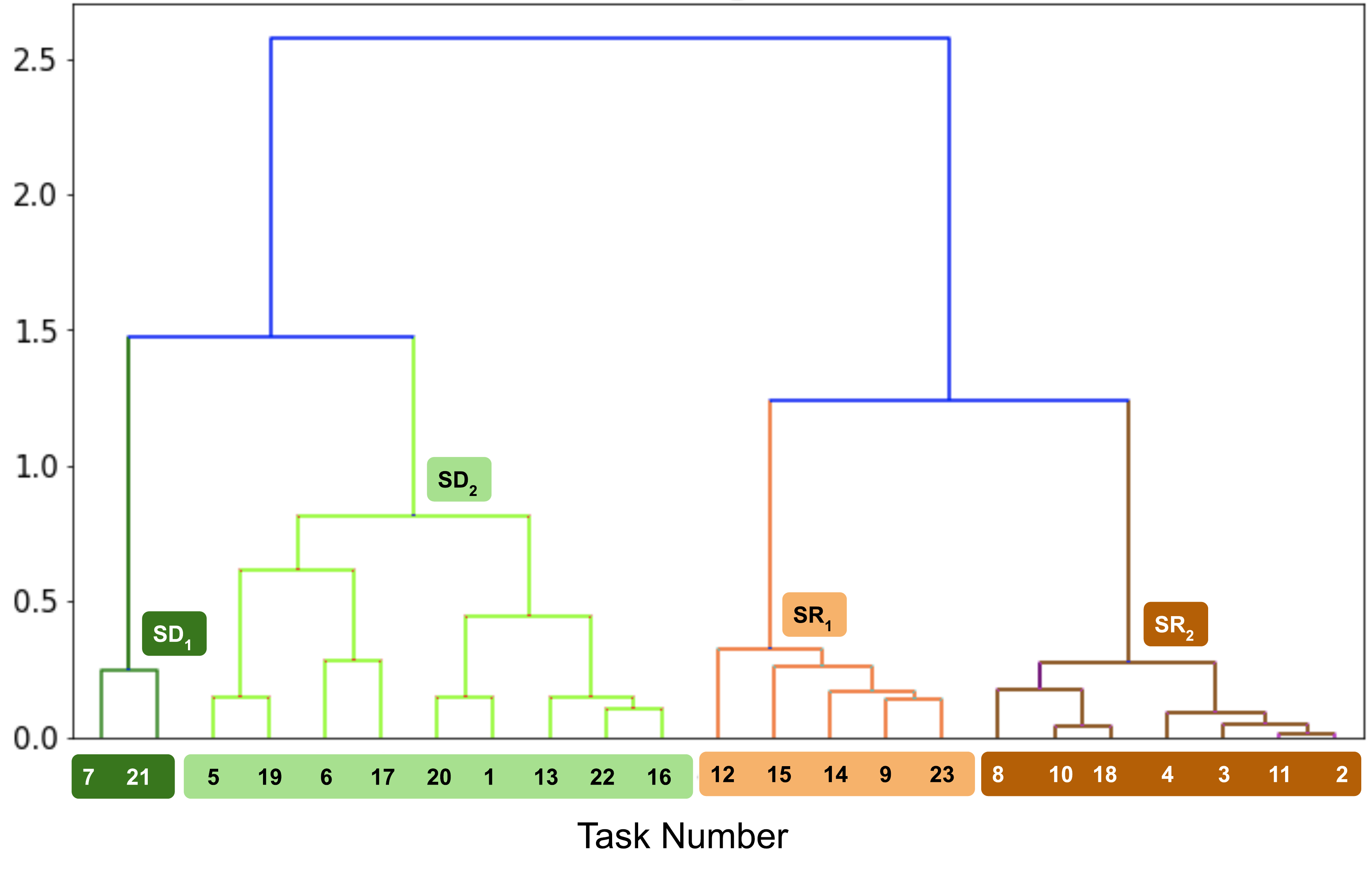}
\caption{Dendrogram derived from an N-dim hierarchical clustering analysis on the test accuracy of N=15 ResNets[18/50/152] trained to solve each task  over a range of training set sizes.} \label{fig:clustering}
\end{figure}

To better understand the computational strategies needed to solve the SVRT, we analyzed ResNet performance on the tasks with a multi-variate clustering analysis. For each individual task, we created an $N$-dimensional vector by concatenating the test accuracy of all ResNet architectures ($N = 3$ depths $\times$ 5 training set sizes = 15), which served as a signature of each task's computational requirements.  We then passed a matrix of these vectors to an agglomerative hierarchical clustering analysis (Figure~\ref{fig:clustering}) using $Ward's$ method.

Our clustering analysis revealed a novel taxonomy for the SVRT.
At the coarsest level, it recapitulated the dichotomy between \textit{same-different} (SD; green branches) and \textit{spatial-relation} (SR; \textcolor{black}{brown} branches) categorization tasks originally identified by \citet{kim2018not} using shallow CNNs. Interestingly, two of the tasks which were classified as SR by \citet{kim2018not} (tasks \textit{6 \& 17}) were assigned to the SD cluster in our analysis. We examined the descriptions of these two tasks as given in~\citet{fleuret2011comparing} (see also Figures~\ref{fig:exampleSD} and~\ref{fig:exampleSR}) and found that these two tasks involve both SR and SD: they ask observers to tell whether shapes are the same or different and judge the distance between the shapes. Specifically, task \textit{6} involves two pairs of identical shapes with one category having same distance in-between two identical shapes vs. not in the other. Similarly in task \textit{17}, three of the four shapes are identical and their distance with the non identical one is same in one category vs. different in the other. Thus, our data-driven dichotomization of SR vs. SD refines the original proposal of \citet{kim2018not}. This could be due to our use of ResNets (as opposed to vanilla CNNs), deeper networks, and a greater variety of training set sizes (including much smaller training set sizes than those used by \citet{kim2018not}). \textcolor{black}{The analysis by \citet{fleuret2011comparing} also revealed that several SD tasks (\textit{6, 16, 17, 21}) are particularly challenging for human observers.}

Our clustering analysis also revealed a finer organization than the main SR vs. SD dichotomy. The SR cluster could be further subdivided into two sub-clusters. The $SR_2$ (dark-brown-coloured) branch in Figure~\ref{fig:clustering} captures tasks that involve relatively simple and basic relation rules such as shapes making close contact (\textit{3, 11}), or being close to one another (\textit{2}), one shape being inside the other (\textit{4}) or whether the shapes are arranged to form a symmetric pattern (\textit{8, 10, 18}). In contrast, tasks that fall in the $SR_1$ (light-brown-coloured) branch involve the composition of more than two rules such as comparing the size of multiple shapes to identify a subgroup before identifying the relationship between the members of the sub-groups. This includes tasks such as finding a \textit{larger} object \textit{in between} two smaller ones (\textit{9}) or three shapes of which two are small and one large with two smaller (\textit{identification of large and small object}) ones either inside or outside in one category vs. one \textit{inside} and the other \textit{outside} in the second (\textit{23}), or \textit{two small} shapes \textit{equally close} to a bigger one (\textit{12}), etc. These tasks also tend to be comparatively harder to learn, requiring ResNets with greater processing depth and more training samples. For instance, tasks \textit{9, 12, 15, 23} were harder to learn than tasks \textit{2, 4, 11} requiring more samples and/or more depth to solve well (Figure~\ref{fig:overall}).

We found that task \textit{15} gets assigned to this latter sub-cluster because the task requires finding four shapes in an image that are identical vs. not. One would expect this task to fall in the SD cluster but we speculate that the deep networks are actually able to leverage a shortcut \citep{geirhos2020shortcut} by classifying the overall pattern as symmetric/square (when the four shapes are identical) vs. trapezoid (when the four shapes are different; see Figure~\ref{fig:exampleSR}) -- effectively turning an SD task into an SR task.

Our clustering analysis also reveals a further subdivision of the SD cluster. These tasks require recognizing shapes which are identical to at least one of the other shapes in the image. The first sub-cluster $SD_2$ (light green color branch) belongs to tasks that require identification of simple task rules, like answering whether or not two shapes are identical (even if it is along the perpendicular bisector) (tasks \textit{1, 20}; see Figure~\ref{fig:exampleSD}), determining if all the shapes on an image are the same (\textit{16, 22}), or detecting if two pairs of identical shapes can be translated to become identical to each other
(\textit{13}). Another set of tasks within this sub-cluster includes tasks that are defined by more complex rules that involve the composition of additional relational judgments. Sample tasks include identifying pairs/triplets of identical shapes and measuring the distance with the rest (\textit{6, 17}), determining if an image consists of pairs of identical shapes (\textit{5}), or detecting if one of the shapes is a scaled version of the other (\textit{19}). Finally, the second sub-cluster $SD_1$ shown in dark-green color involves two tasks that require an understanding of shape transformations. One task asks observers to say if one of the shapes is the scaled, translated, or rotated version of the other one (\textit{21}). The other task tests asks observers to judge whether or not an image contains two pairs of three identical shapes or three pairs of two identical shapes in an image (\textit{7}).  

To summarize this first set of experiments, we have systematically evaluated the ability of ResNets spanning multiple depths to solve each of the twenty-three SVRT tasks for different training set sizes. This allowed us to represent SVRT tasks with according to their learnability by ResNets of varying depth. By clustering these representations, we extracted a novel SVRT taxonomy that both recaptulated an already described SD-SR dichotomy \citep{kim2018not}, and also revealed a more granular task structure corresponding to the number of rules used to form each task. Tasks with more rules are harder for ResNets to learn. Our taxonomy also reveals an organization of tasks where easier $SR_1$ and $SR_2$ sub-clusters fall closer to each other than harder $SD_1$ and $SD_2$ sub-clusters.

\section*{Experiment 2: Attention networks}
\label{sec:exp2}
We next sought to identify computational mechanisms that could help ResNets learn the more challenging SVRT tasks revealed by our novel taxonomy. Attention has classically been implicated in visual reasoning in primates and humans \citep{egly1994covert,roelfsema1998object}. Attentional processes can be broadly divided into \textit{\textbf{spatial}} (e.g., attending to all features in a particular image location) vs. \textit{\textbf{feature-based}} (e.g., attending to a particular shape or color at all spatial positions) \citep{desimone1995neural}. The importance of attention for perceiving and reasoning about challenging visual stimuli has also been realized by the computer vision community. There are now a number of attention modules proposed to extend CNNs -- including spatial  (e.g., \citet{Sharma2015-ow,journals/corr/ChenWCGXN15, yang2016, journals/corr/XuS15a,journals/corr/RenZ16}), feature-based (e.g., \citet{stollenga2014deep, chen2017sca, hu2018squeeze}) and hybrid (e.g., \citet{linsley2018global,woo2018cbam}) approaches. Here, we adapt the increasingly popular Transformer architecture \citep{vaswani2017transformer} to implement both forms of attention. These networks, which were originally developed for natural language processing, are now pushing the state of the art in computer vision \citep{Zhu2020-nc,Carion2020-dw,dosovitskiy2020image}. \textcolor{black}{Recent work \citep{Ding2020AttentionOL} has also showed the benefits of such architectures and especially attention mechanisms for solving higher-level reasoning problems}.

\textcolor{black}{Transformers are neural network modules which usually consist of at least one ``self-attention'' module followed by a feedforward layer. Here, we introduced different versions of the self-attention module into ResNets to better understand the computational demands of each SVRT task. The self-attention implemented by Transformers is both applied to and derived from the module's input. By reconfiguring standard Transformer self-attention, we developed versions capable of allocating either spatial or feature-based attention over the input. Specifically, we created these different forms of attention by reshaping the convolutional feature map input to a transformer. For spatial attention, we reshaped the $ \mathcal{Z} \in \mathcal{R}^{H, W, C}$ feature maps to $ \mathcal{Z} \in \mathcal{R}^{C, H * W}$, so that the Transformer's self-attention was allocated over all spatial locations. For feature-based attention, we reshaped the convolutional feature maps to $\mathcal{Z} \in \mathcal{R}^{H * W, C}$, enforcing attention over all features instead of spatial locations. See section~\ref{SI:1} for an elaborated treatment.}

\begin{figure}[h]
    \centering
    \includegraphics[width=.9\linewidth]{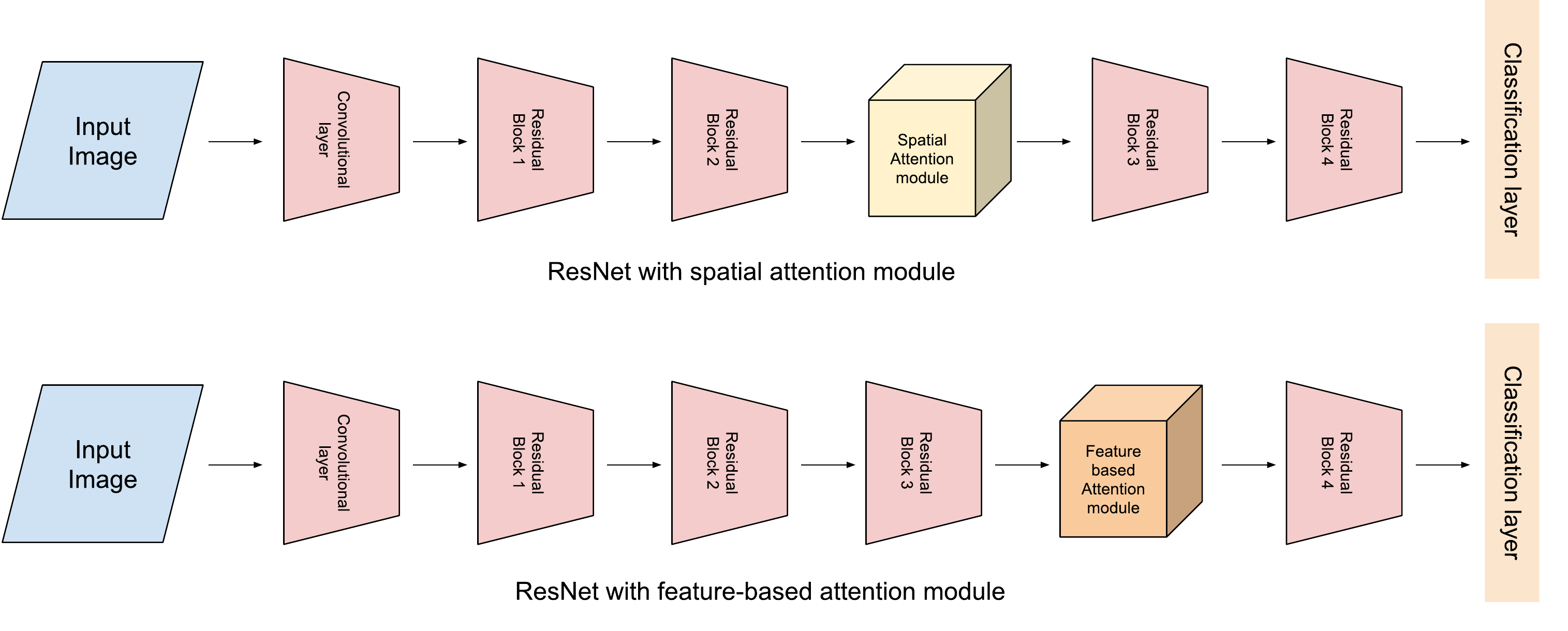} 
    \caption{\label{fig:architecture} Location of the Transformer self-attention modules in our ResNet extensions.}
    
\end{figure}

\textcolor{black}{ We added one spatial or feature-based attention after one of the four residual blocks in a ResNet-50. We selected where to add either form of attention to a ResNet-50 by choosing the location where the addition of attention yielded the best validation accuracy across the SVRT tasks. Through this procedure, we inserted a spatial attention module after the second residual block, and a feature-based attention module after the third residual block (Figure~\ref{fig:architecture}).}

To measure the effectiveness of different forms of attention for solving the SVRT, we compared the accuracy of three ResNet-50 models: one capable of spatial attention, one capable of feature-based attention, and one that had no attention mechanisms (``vanilla''). Spatial attention consistently improved model accuracy on all tasks, across all five dataset sizes that models we used for training models. The improvement in accuracy is particularly noticeable for the $SD_1$ cluster. Tasks in this sub-cluster are composed of two rules, which ResNets without attention struggled to learn. Attention helps ResNets learn these tasks more efficiently. The improvement is also evident for $SD_2$ and $SR_1$. The benefit of attention for $SR_2$ is however marginal, since ResNets without attention already perform well on these tasks. 

\begin{figure}[t]
\centering
 \includegraphics[width=1\linewidth]{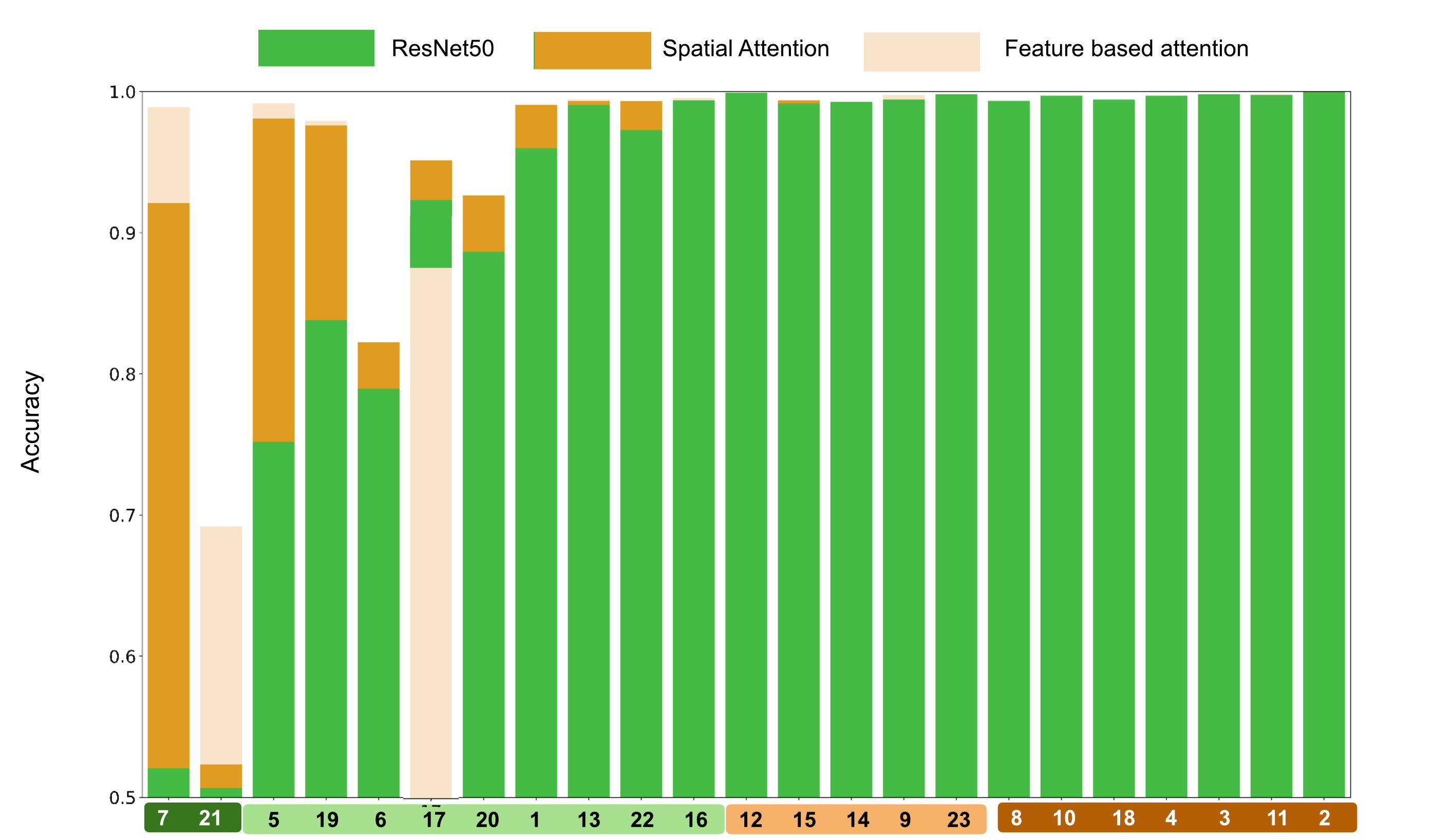}
 \caption{Test accuracies for 50-layer ResNets with spatial attention (orange), feature-based attention (tan), or no attention (green). Each bar depicts performance after training from scratch on 10k samples. }
 \label{fig:fbsa}
\end{figure}

We find that feature-based attention leads to the largest improvements for $SD_1$, especially when training on 5k or 10k examples 
(Figure~\ref{fig:fbsa}). On the other hand, spatial attention leads to the largest improvements for $SD_2$ and $SR_1$. This improvement is pronounced when training on 500 or 1000 examples. Taken together, the differential success of spatial versus feature-based attention reveal that the task sub-clusters discovered in our data-driven taxonomy can be explained by their varying attentional demands.

\begin{figure*}[ht!]
\centering
  \begin{subfigure}{.8\linewidth}
    \centering
    \includegraphics[width=.9\linewidth]{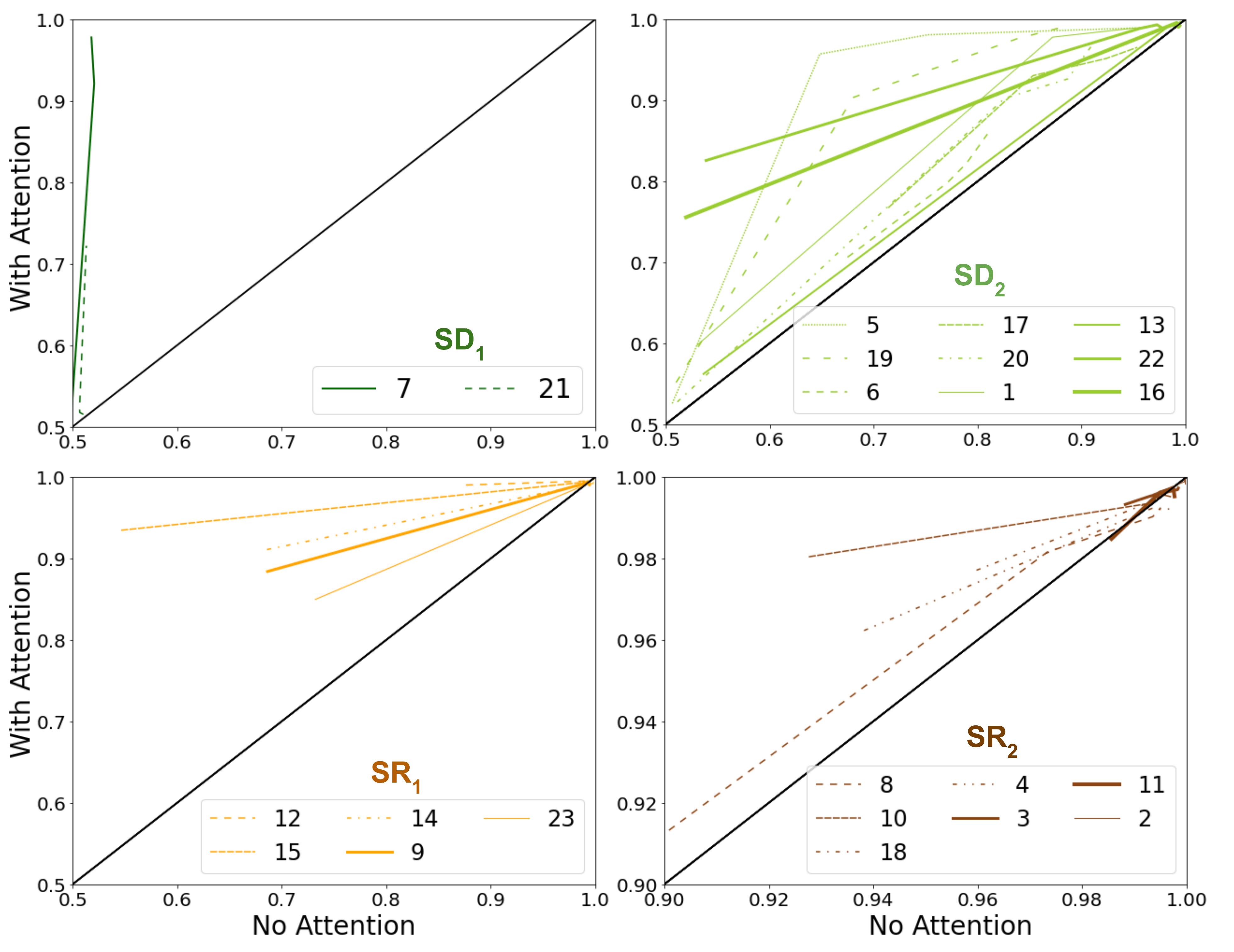} 
    \subcaption{Spatial attention}
  \end{subfigure}
  
  \begin{subfigure}{.8\linewidth}
    \centering
    \includegraphics[width=.9\linewidth]{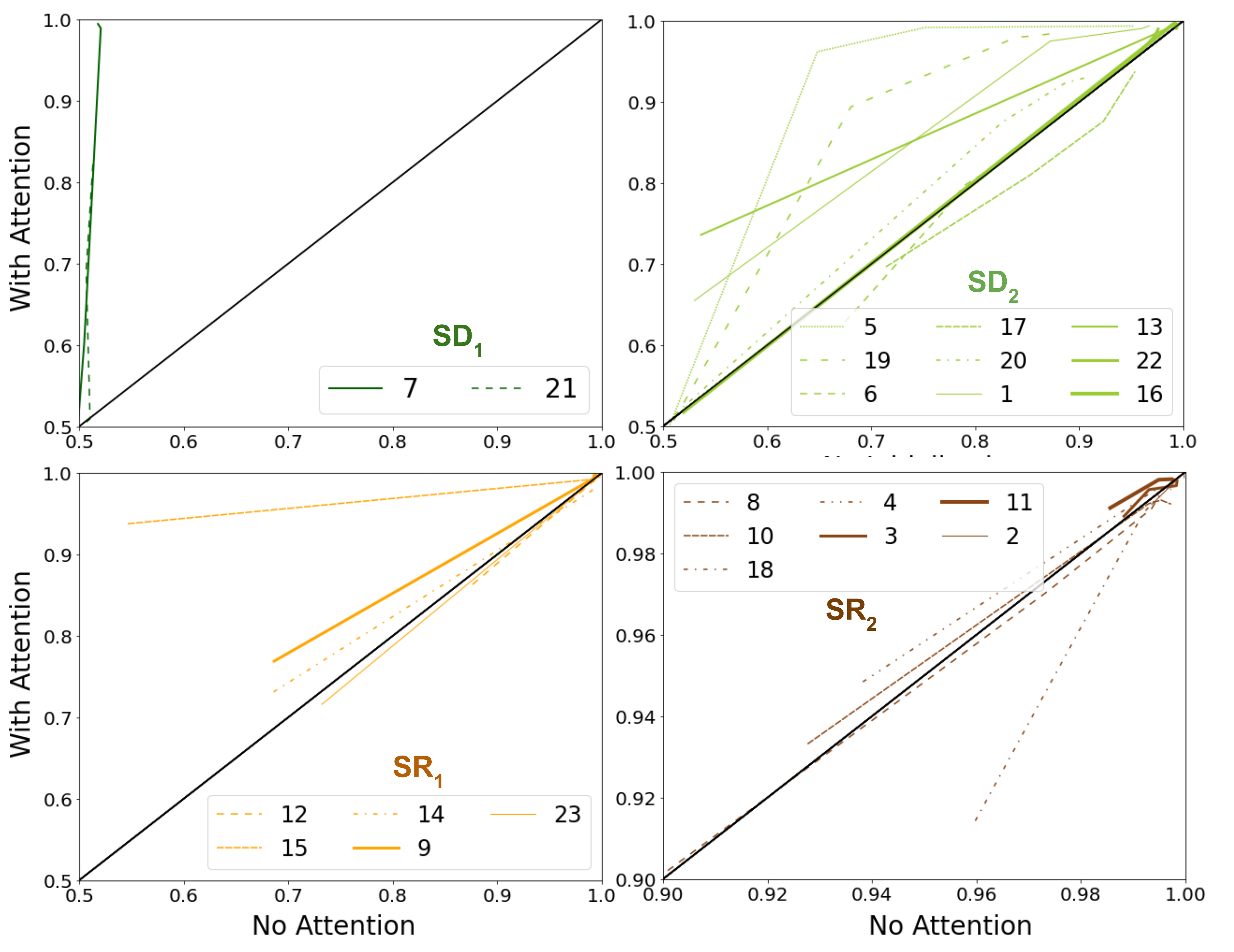}
  \subcaption{Feature-based attention}
 \end{subfigure}%
 \caption{Test accuracies for a baseline ResNet50 vs. the same architecture endowed with the two forms of attention for each of the twenty-three SVRT tasks when varying the number of training examples. A different axis scale is used for $SR_2$ to improve visibility. \textcolor{black}{These curves are constructed by joining task accuracy for five points representing dataset sizes.} }\label{fig:xy_sa}
\end{figure*}

\clearpage

\begin{figure*}[ht!]
\centering
  \includegraphics[width=1\linewidth]{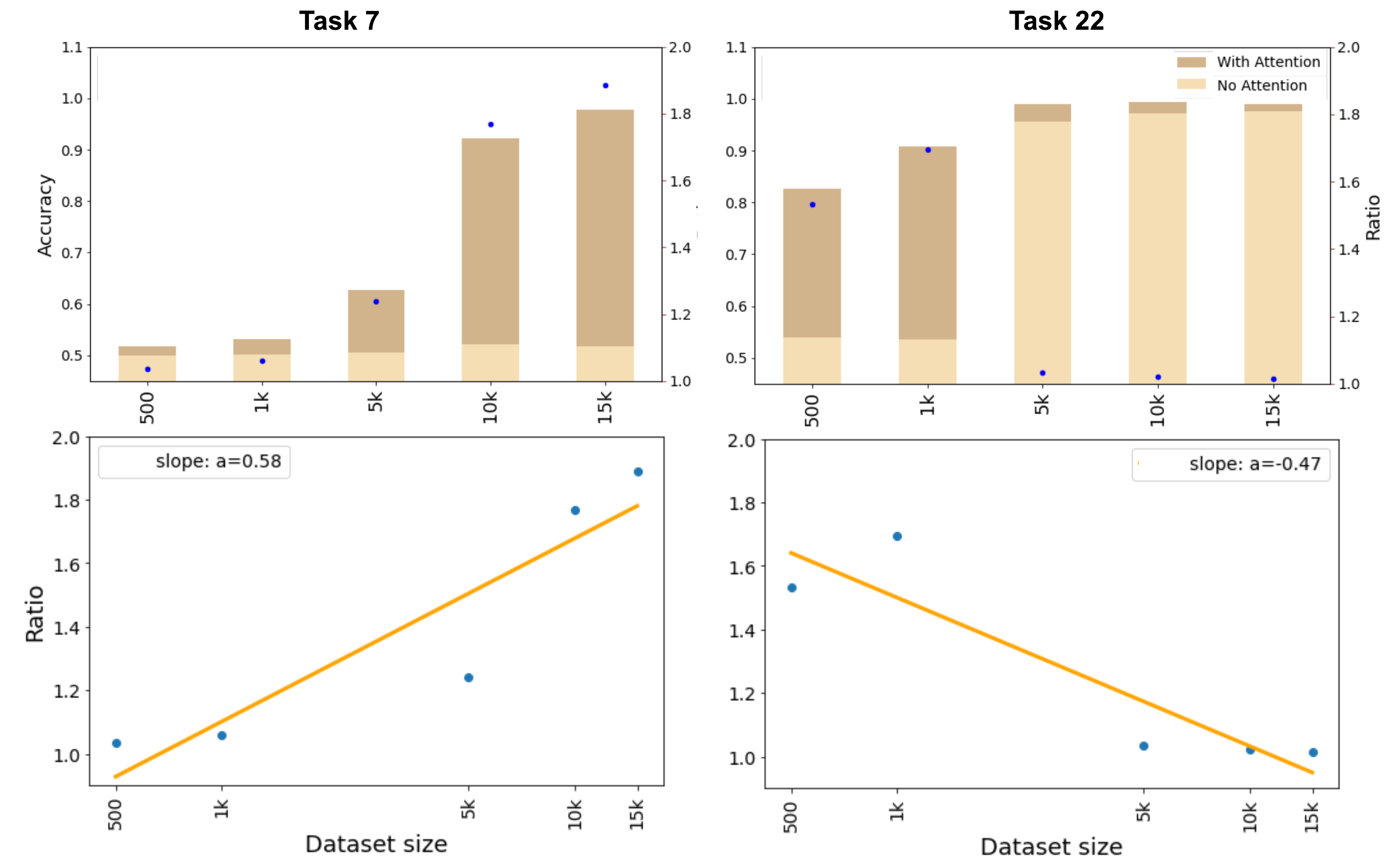}
 \caption{The benefit of attention in solving the SVRT is greatest in data-limited training regimes. The x-axis depicts the number of samples for training, and the y-axis depicts a ratio of the average performance of models with attention to models without attention. When the ratio is greater than 1 it shows that attention helps vs. hurts when lower than 1. This gives us five ratios per task and attention process corresponding to each dataset size. We performed a linear fitting procedure for these points and calculated the corresponding slope. This slope characterizes the relative benefits of attention for that particular task as the number of training examples available increases. If the benefit of attention is most evident in lower training regimes, one would expect a relatively small slope and if the benefit of attention is most evident in higher training regimes, one would expect a large slope.
 }\label{fig:slope}
\end{figure*}

To better understand how the ResNet-derived taxonomy found in Experiment 1 can be explained by the need for spatial and feature-based attention, we measured the relative improvement of each form of attention over the vanilla ResNet. For each attention model and task, we calculated the ratio of the test accuracies between the model and the vanilla ResNet50. We repeated this for every training dataset size, then fit a linear model to these ratios to calculate the slope across dataset sizes (see Figure~\ref{fig:slope} for representative examples). We then repeated this procedure for all twenty-three tasks to produce two 23-dimensional vectors containing slopes for each model and every task.

\begin{figure*}[t]
\centering
  \includegraphics[width=.8\linewidth]{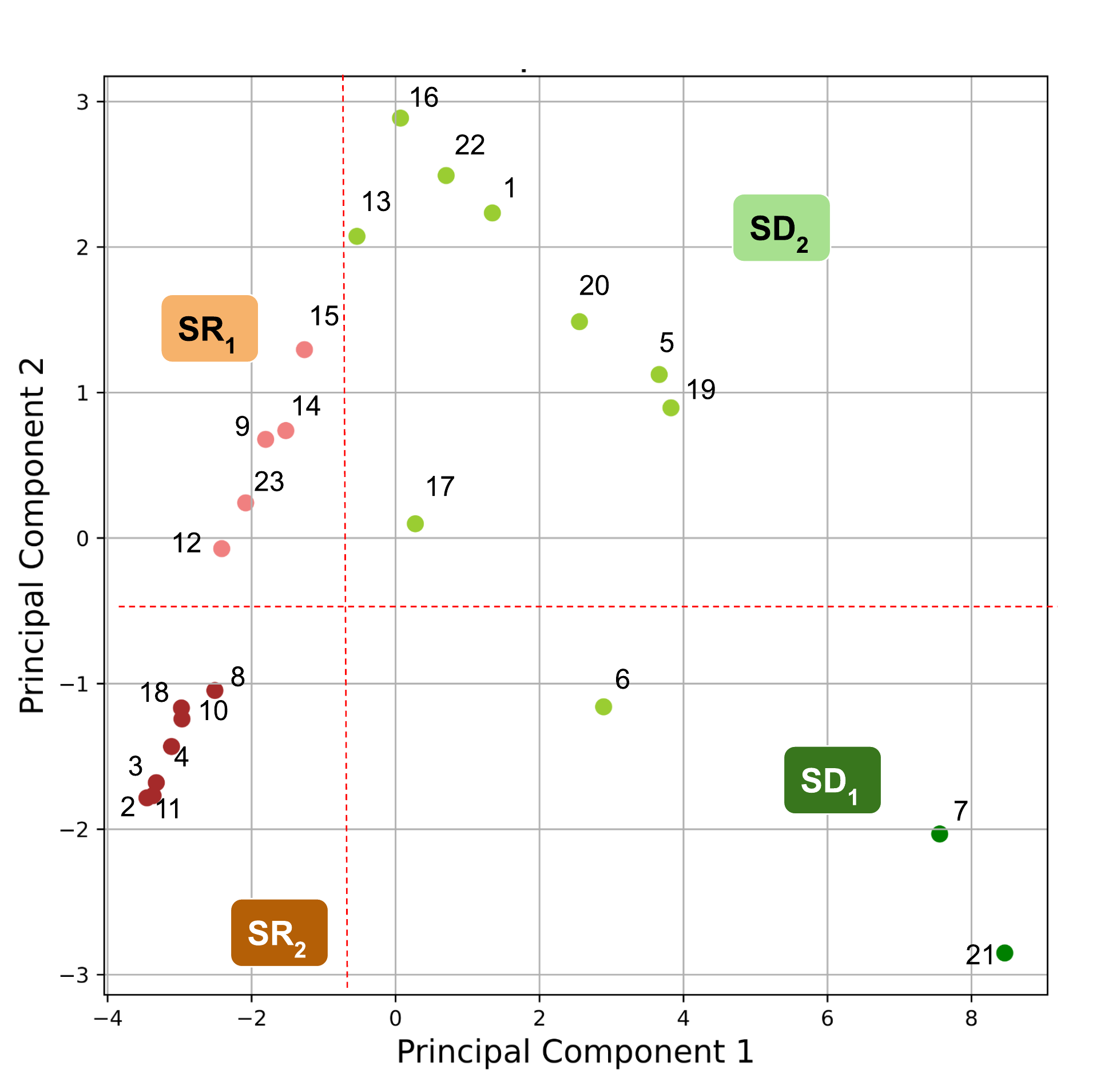}
 \caption{Principal component analysis of the twenty-three tasks using the 15-dimensional feature vectors derived from Experiment 1 representing the test accuracy obtained for each task for different dataset sizes and ResNets of varying depths (18, 50 \& 152). Dotted red line represents 4 different bins in which these tasks can be clustered.} \label{fig:pca}
\end{figure*}

We next used these slopes to understand the attentional demands of each SVRT task. We did this through a two-step procedure. First, we applied a principal component analysis (see Figure~\ref{fig:pca}) to the vanilla ResNet performance feature vectors ($N=15$) derived from Experiment 1. Second, we correlated the principle components with the slope vectors from the two attention models. We restricted our analysis to the first two principle components, which captured  $\sim 93\%$ of the variance in the vanilla ResNet's performance (Figure~\ref{fig:pca}). This analysis revealed a dissociation between the two forms of attention: feature-based attention was most correlated with the first principal component and spatial attention with the second principal component. Additionally, along the first principal component, we found the broader dichotomy of these 23 tasks into $SD$ and $SR$ clusters, whereas the second principal component divulge the tasks which responded better with spatial attention from tasks requiring either no attention or feature based attention (as seen in dotted red line along both the axis in Figure~\ref{fig:pca}).  The corresponding Pearson coefficient $r$ and $p$ values are given in Table~\ref{tab:rp}. 

\begin{table}[h]
\centering
\begin{center}
    \caption{Pearson coefficient ($r$) and corresponding $p$ values obtained by correlating the slope vectors of the spatial attention and the feature-based attention modules with the two principal components of Figure~\ref{fig:pca}. See text for details.}
    \label{tab:rp}

\end{center}

\begin{tabular}{lcccc} 
\hline 
   & \multicolumn{2}{c}{$Spatial$} & \multicolumn{2}{c}{$Feature$} \\ \hline 
   & \textbf{r }      & \textbf{p }     & \textbf{r }         & \textbf{p }        \\  \hline \hline 
$PC_{1}$ & 0.466                 & 0.0249                & \textbf{0.649}            & 0.0008                  \\ \hline
$PC_{2}$ & \textbf{-0.652}       & 0.0007               & -0.491                   & 0.0174 \\ \hline
\end{tabular}

\end{table}

To summarize our results from Experiment 2,  we have found that the task clusters derived from ResNet test accuracies computed over a range of depth and training set sizes can be explained in terms of attentional demands. Here, we have shown that endowing these networks with attentional mechanisms helps them learn some of the most challenging problems with far fewer training examples. We also found that the relative improvements obtained over standard ResNets with feature-based and spatial attention are consistent with the taxonomy of visual reasoning tasks found in Experiment 1. More generally, our analysis shows how the relative need for feature vs. spatial attention seems to account for a large fraction of the variance in computational demand required for these SVRT tasks defined in Experiment 1 according to their learnability by ResNets. 

\section*{Experiment 3: Feature vs. rule learning}

The learnability of individual SVRT tasks reflects two components: the complexity of the task's visual features, and separately, the complexity of the rule needed to solve the task. To what extent are our estimates of learnability driven by either of these components? We tested this question by training a new set of ResNets without attention according to the procedure laid out in Experiment 1, but with a different pre-training strategies. One of the ResNets was pre-trained to learn visual statistics (but not rules) of SVRT images, and another was pre-trained on ImageNet, \citep[a popular computer vision dataset containing natural object categories;][]{deng2009imagenet}.

For pre-training on SVRT, we sampled 5,000 class-balanced images from each of the 23 tasks (5,000 $\times$ 23 = 115,000 samples in total). To make sure the networks did not learn any of the SVRT task rules, we shuffled images and binary class labels across all twenty-three problems while pre-training the network. We then trained models with binary crossentropy to detect positive examples \textit{without discriminating tasks}. \textcolor{black}{Our assumption is that shuffling images and labels removes any semantic information between individual images and SVRT rules. However, a network with sufficient capacity can still learn the corresponding mapping between arbitrary images and class labels (even though it cannot generalize it to novel samples). To learn this arbitrary mapping, the network has to be able to encode visual features; but by construction, it cannot learn the SVRT task rule.} When training this model and the ImageNet initialized model to solve individual SVRT tasks, we froze the weights of the convolutional layers and only fine-tuned the classification layers to solve SVRT problems.

\begin{figure*}[htbp]
\centering
    \includegraphics[width=1\linewidth]{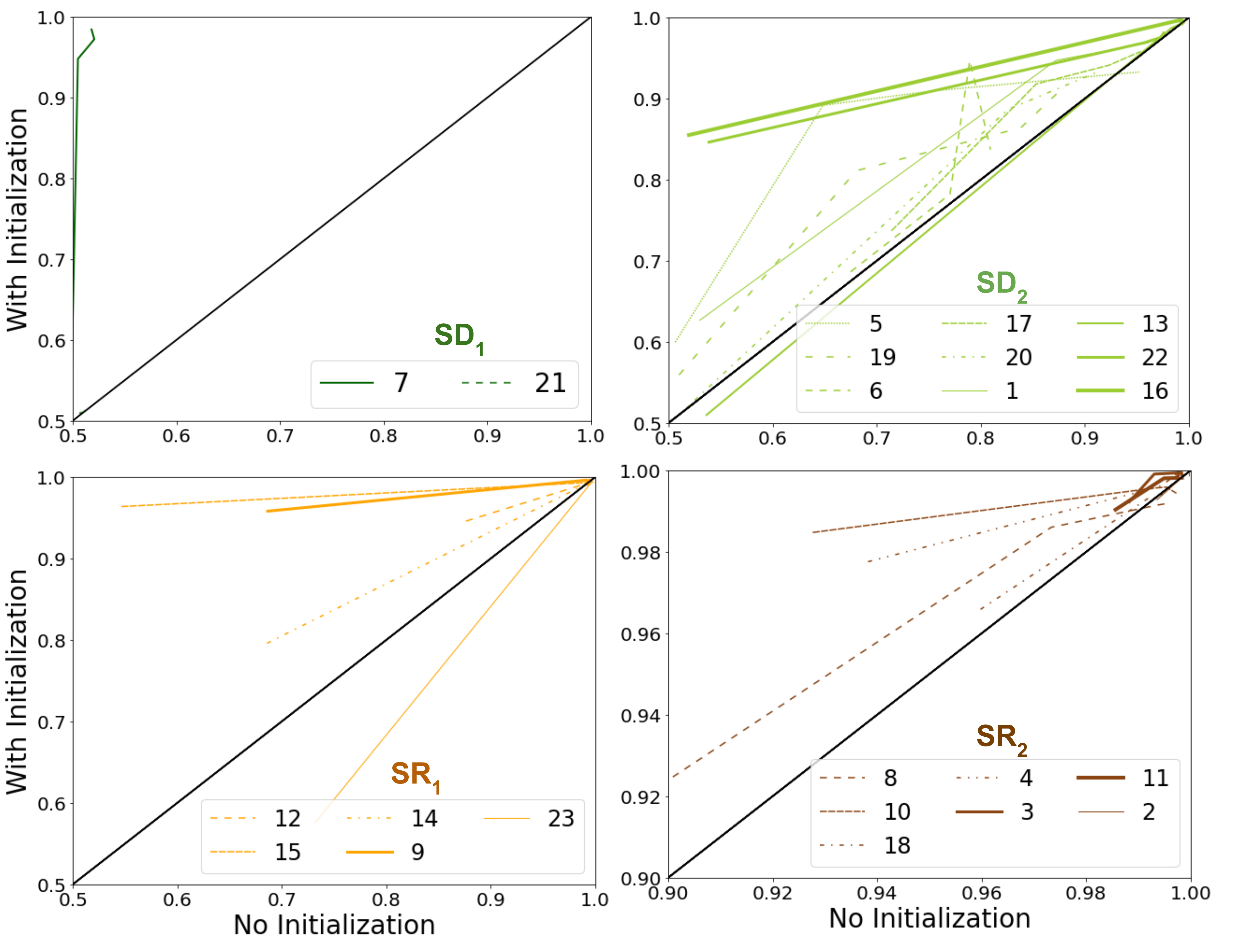}
 \caption{Test accuracies for a baseline ResNet50 trained from scratch (``No initialization'') vs. the same architecture pre-trained on an auxiliary task in order to learn visual representations that are already adapted to the SVRT stimuli for different \textcolor{black}{numbers} of training examples. The format is the same as used in Figure \ref{fig:xy_sa}. A different axis scale is used for $SR_2$ to improve visibility. \textcolor{black}{These curves are constructed by joining task accuracy for five points representing dataset sizes.}
 }\label{fig:xy_c_f}
\end{figure*}

Figure~\ref{fig:xy_c_f} shows a comparison between the different architectures in terms of their test accuracies according to the sub-clusters discovered in Experiment 1. These results first confirm that the SVRT pre-training approach works because it consistently outperforms pre-training on ImageNet (Figure~\ref{fig:xy_img}) or training from scratch. Interestingly, for the $SR_{2}$ sub-cluster, we found that the benefits of pre-training on SVRT goes down very quickly as the number of training examples grows. We interpret these results as reflecting the fact that generic visual features are sufficient for the task and that the rule can be learned very quickly (somewhere around 500 and 5,000 samples). For $SR_{1}$ sub-cluster, the benefits of starting from features learned from SVRT is somewhat more \textcolor{black}{evident} in low training regimes but these advantages quickly vanish as more training examples are available (the task is learned by all architectures within 5,000 training samples). 

For $SD_{1}$ while there appears to be a \textcolor{black}{noteworthy} advantage of pre-training on SVRT over ImageNet pre-training and training from scratch, the tasks never appear to be fully learned by any of the networks even with 15,000 training examples. This demonstrates the challenge of learning the rules associated with this sub-cluster beyond simply learning good visual representations. Finally, our results also show that the performance gap across all the architectures for $SD_{2}$ vs. $SD_{1}$ increases rapidly with more training examples -- demonstrating the fact that the abstract rule for $SD_{2}$ tasks are more rapidly learned than for $SD_{1}$.

\textcolor{black}{Finally}, we carried out a similar analysis with the pre-trained network as done in Experiment 2: We built test accuracy vectors for the SVRT-pre-trained network trained using all five dataset sizes (.5k, 1k, 5k, 10k, 15k) and searching over a range of optimal learning rates (\textit{1e-4, 1e-5, 1e-6}). This led to five-dimensional vector, which we  normalize by dividing each entry with the corresponding test accuracy of a baseline ResNet50 trained from scratch. Hence, the normalized vector represent the improvement (ratio larger than 1) or reduction in accuracy (ratio smaller than 1) that results from the pre-training on SVRT for that particular task and training set size. We then calculated the slope vector in $\mathcal R^{(23)}$, which we correlated with the corresponding spatial and feature-based attention vectors from Experiment 2.

\textcolor{black}{We found that task improvements due to SVRT pre-training correlated more strongly with task improvements due to spatial ($r=0.90$, $p=4e-9$) than feature-based attention ($r=0.595$, $p=0.002$)}. This suggests that the observed improvements in accuracy derived from spatial attention are more consistent with learning better feature representations compared to feature-based attention. 

To summarize, in Experiment 3, we have tried to address the question of learnability of SVRT features vs. rules. We found that  using an auxiliary task to pre-train the networks on the SVRT stimuli in order to learn visual representations beforehand provides learning advantages to the network compared to a network trained from scratch. 

We also found a noteworthy correlation between the test accuracy vector of a network pre-trained on SVRT visual statistics and a similar network endowed with spatial attention. This suggests that spatial attention helps discover the abstract rule more so that it helps improve learning good visual representations for the task. 

\section*{\textcolor{black}{Conclusion}}

The goal of the present study was to shed light on the computational mechanisms underlying visual reasoning using the Synthetic Visual Reasoning Test (SVRT)~\citep{fleuret2011comparing}. \textcolor{black}{There are} twenty-three binary classification problems in this challenge, which include a variety of same-different and spatial reasoning tasks.  

In a first experiment, we systematically evaluated the ability of a battery of $N=15$ deep convolutional neural networks (ResNets) -- varying in depths and trained using different training set sizes -- to solve each of the SVRT problems. We found a range of accuracies across all twenty-three tasks, with some tasks being easily learned by shallower networks and relatively small training sets, and some tasks remaining hardly solved with much deeper networks and orders of magnitude more training examples.

Under the assumption that the computational complexity of individual tasks can be well characterized by the pattern of test accuracy across these $N=15$ neural networks, we formed N-dimensional accuracy vectors for each task and ran a hierarchical clustering algorithm. The resulting analysis \textcolor{black}{suggests} a taxonomy of visual reasoning tasks: beyond two primary clusters corresponding to same-different (SD) vs. spatial relation (SR) judgments, we also identified a finer organization with sub-clusters reflecting the nature and the number of relations used to compose the rules defining the task. Our results are consistent with previous work by \citet{kim2018not} who first identified a dichotomy between SD and SR tasks. Our results also extend prior work \citep{fleuret2011comparing,kim2018not,yihe2019program} in \textcolor{black}{proposing} a finer-level taxonomy of visual reasoning tasks. The accuracy of neural networks is reflected in the number of relationships used to define the basic rules, which is expected, but it deserves closer examination. 

\citet{kim2018not} have previously suggested that SD tasks ``strain'' convolutional neural networks. That is, while it is possible to find a network architecture of sufficient depth (or number of units) that can solve a version of the task up to a number of stimulus configurations (e.g., by forcing all stimuli to be contained within a $\Delta H \times \Delta W$ window), it is relatively easy to render the same task unlearnable by the same network \textcolor{black}{past} a certain number of stimulus configurations (e.g., by increasing the size of the window that contains all stimuli). It is as if these convolutional networks are \textcolor{black}{capable of learning} the task if the number of stimulus configurations remains below their memory capacity, and fail beyond that. \textcolor{black}{It remains an open question  whether non-convolution} alternatives to the CNNs tested here such as the now popular transformer networks  \citep{dosovitskiy2020image,touvron2021training,tolstikhin2021mlp} would learn to solve some of the harder SVRT tasks more efficiently. As an initial experiment, we attempted to train and test a Vision Transformer \footnote[1]{\url{https://github.com/facebookresearch/dino}} (ViT) \citep{dosovitskiy2020image}  constrained to have a similar number of parameters (21M) to the ResNet-50 used here. We were not able to get these architectures to do well on most of the tasks that are difficult for ResNets even with 100k samples (also shown in \citet{messina2021recurrent}). It is worth noting that even 100k samples remain a relatively small dataset size by modern day standards since ViT was trained from scratch.

Multi-layer perceptrons and convolutional neural networks including ResNets an other architectures can be formally shown to be universal approximators under certain architectural constraints. That is, they can learn arbitrary mappings between images to class labels. Depending on the complexity of the mapping one might need an increasing number of hidden units to allow for enough expressiveness of the network; but provided enough units $/$ depth and a sufficient amount of  training examples, deep CNNs can learn arbitrary visual reasoning tasks. While we cannot make any strong claim for the specific ResNet architectures used in this study (currently the proof is limited to single layer without max pooling or batch normalization \citep{lin2018resnet}), we have indeed found empirically that all SVRT tasks could indeed be learned for networks of sufficient depth and provided a sufficient amount of training examples.
However, deep CNNs typically lack many of the human cognitive functions such as attention and working memory. Such functions are likely to provide a critical advantage for a learner to solve some of these tasks \citep{10.7551/mitpress/1187.001.0001}. CNNs might have to rely instead on function approximation which could lead to a less general ``brute-force'' solution. 
Given this, an open question is whether the clustering of SVRT tasks derived from our CNN-based analyses will indeed hold for human studies. At the same time, the prediction by \citet{kim2018not} using CNNs that SD tasks are harder than SR tasks and hence that they may demand additional computations (through feedback processes) such as attention and/or working memory was successfully validated experimentally by by \citet{AlamiaENEURO.0267-20.2020} using EEG. 

Additional evidence for the benefits of feedback mechanisms for visual reasoning was provided by  \citet{linsley2018learning} who showed that contour tracing tasks that can be solved efficiently with a single layer of a recurrent-CNN may require several order of magnitudes more processing stages in a non-recurrent-CNN to solve the same task. This ultimately translates into much greater sample efficiency for recurrent-CNNs on natural image segmentation tasks \citep{linsley2020recurrent}. \textcolor{black}{The closely related task of ``insideness'' was also studied by \citet{villalobos2021neural} who demonstrated the inability of CNNs to learn a general solution for this class of problems.}
Universal approximators with minimal inductive biases such as multi-layer perceptrons, CNNs and other feedforward or \textcolor{black}{non}-attentive architectures can learn to solve visual reasoning tasks, but they might need a very large number of training examples to properly fit. Hence, beyond simply measuring the accuracy of very deep nets in high data regimes (such as when millions of training examples are available), systematically assessing the performance of neural nets of varying depths and for different training regimes may provide critical information about the complexity of different visual reasoning tasks. 

\citet{kim2018not} hypothesized that such straining by convolutional networks is due to their lack of attention mechanisms to allow the explicit binding of image regions to mental objects. A similar point was made by \citet{greff2020binding} in the context of the contemporary neural network failure to carve out sensory information into discrete chunks which can then be individually analyzed and compared (see also \citet{10.1007/978-3-540-75555-5_15} for a similar point). Interestingly, this prediction was recently tested using human EEG by \citet{AlamiaENEURO.0267-20.2020} who showed that indeed the brain activity recorded during SD tasks is compatible with greater attention and working memory demands than SR tasks.

\textcolor{black}{At the same time, that CNNs can learn SR tasks more efficiently than SD tasks does not necessarily mean that human participants can solve these tasks without attention. Indeed, \citep{logan1994spatial} has shown that under some circumstances SR tasks such judging insideness requires attention.} 

To further assess the role of attention in visual reasoning, we used transformer modules to endow deep CNNs with spatial and feature-based attention. The relative improvements obtained by the CNNs with the two forms of attention varied across tasks. Many tasks reflected a larger improvement for spatial attention, and a smaller number benefited from feature-based attention. Further, we found that the patterns of relative improvements accounted for much of the variance in the space of SVRT tasks derived in Experiment 1. Overall, we found that the requirement for feature-based and spatial attention accounts well for the taxonomy of visual reasoning tasks identified in Experiment 1.  \textcolor{black}{Our} computational analysis also lead to testable predictions for human experiments by suggesting tasks that either benefit from spatial attention (task \textit{22}) or from feature-based attention (task \textit{21}), tasks that benefit from either form of attention (task \textit{19}), and tasks that do not benefit from attention (task \textit{2}).

Finally, our study has focused on the computational benefits of spatial and feature-based attention for visual reasoning. Future work should consider the role of other forms of attention including object-based attention \citep{egly1994covert} for visual reasoning.

In our third experiment, we studied the learnability of SVRT features vs. rules. We did this by pre-training the neural networks on auxiliary tasks in order to learn SVRT features before training them to learn the abstract rules associated with individual SVRT problems. Our pre-training methods led to networks that learn to solve the SVRT problems better than networks trained from scratch as well as networks that were pre-trained to perform image categorization on the ImageNet dataset. We have also found that such attention processes seem to contribute more to rule learning than to feature learning. For $SR_1$ sub-cluster we find this type of pre-training to be advantageous in lower training regimes but the benefits rapidly fade away in higher training regimes. In contrast, this pre-training does not allow the tasks from the $SD_1$ sub-cluster to be learned even with 15k samples -- suggesting that the key challenge with these tasks is not to discover good visual representations but rather to discover the rule. This suggests the need for additional mechanisms beyond those implemented in ResNets. This is also consistent with the improvements observed for these tasks with the addition of attention mechanisms.

In summary, our study compared the computational demands of different visual reasoning tasks. While our focus has been on understanding the computational benefits of attention  and feature learning mechanisms, it is clear that additional mechanisms will be required to fully solve all SVRT tasks. These mechanisms are likely to include working memory which is known to play a role in SD tasks \citep{AlamiaENEURO.0267-20.2020}. Overall, this work illustrates the potential benefits of incorporating brain-like mechanisms in modern neural networks and provides a path forward to achieving human-level visual reasoning.

\subsection*{Acknowledgments}
This work was funded by NSF (IIS-1912280) and ONR (N00014-19-1-2029) to TS and ANR (OSCI-DEEP grant ANR-19-NEUC-0004) to RV. Additional support was provided by the ANR-3IA Artificial and Natural Intelligence Toulouse Institute (ANR-19-PI3A-0004), the Center for Computation and Visualization (CCV) and High Performance Computing (HPC) resources from CALMIP (Grant 2016-p20019). We acknowledge the Cloud TPU hardware resources that Google made available via the TensorFlow Research Cloud (TFRC) program as well as computing hardware supported by NIH Office of the Director grant S10OD025181.

\bibliographystyle{apalike}
\bibliography{main}
\newpage

\clearpage

\begin{center}
\textbf{\Huge Supplementary Information}
\end{center}

\renewcommand{\thefigure}{S\arabic{figure}}
\renewcommand{\thesection}{S\arabic{section}}
\setcounter{figure}{0} 
\setcounter{section}{0} 

\section{ResNet50 with attention}
\label{SI:1}

Here we detail how we adapt the attention module originally developed for natural language processing in \citep{vaswani2017transformer} and insert it in an already existing convolutional network architecture such as ResNet50. Similar adaptations can be found in recent works \citep{carion2020end, ramachandran2019stand}.

\paragraph{Spatial Attention Module (SAM)}

Our first attention module takes a features map $X \in \mathcal{R}^{d_C \times d_H \times d_W}$ as input, where $d_C$, $d_H$, and $d_W$ respectively  refer to the number of channels, height and width of the map, and outputs a features map $Y$ of the same dimensions.
We flatten the spatial dimensions to obtain $X' \in \mathcal{R}^{d_C \times d_N}$, where $d_N = d_H \times d_W$, and we apply the original multi-head self-attention module from \citet{vaswani2017transformer} as follows.

We first apply independent linear mappings of the input $X'$ to obtain three features maps of dimensions $\mathcal{R}^{d \times d_N}$ for each attention head from a total of $n_H$ heads. For the $i^{th}$ head, these maps are known as the query $Q_i$, the key $K_i$ and the value $V_i$, and are obtained such as:
\begin{align*}
Q_i = W^Q_i . X' \\
K_i = W^K_i . X' \\
V_i = W^V_i . X'
\end{align*}
The mappings are parametrized by three matrices $W^Q_i$, $W^K_i$ and $W^V_i$ of dimensions $\mathcal{R}^{d \times d_C}$ for each head. The symbol $.$ denotes a matrix multiplication.

Then, we apply the scaled dot-product attention \citep{vaswani2017transformer} to obtain $n_H$ attention heads of dimensions $\mathcal{R}^{d \times d_N}$ such as:
\begin{align}
    H_i = SoftMax(\frac{Q_i . K_i^T}{\sqrt{d}}) V_i
\end{align}

After, we concatenate all attention heads along the first dimension and apply a linear mapping to obtain $Y' \in \mathcal{R}^{d_C \times d_N}$ such as:
\begin{align}
    Z = W^O . Concat(H_1, ..., H_{n_H}) 
\end{align}
The mapping is parametrized by the matrix $W^O \in \mathcal{R}^{d_C \times d}$.

As commonly done, we have a residual connection before applying a layer normalization \citep{ba2016layer} such as:
\begin{align}
    Y' = LayerNorm(Z + X')
\end{align}

Finally, we unflatten $Y'$ to obtain $Y \in \mathcal{R}^{d_C \times d_H \times d_W}$.

We obtain the best results with a representation space of 512 dimensions ($d=512$) and four attention heads ($n_H=4$).

\paragraph{Features-based Attention Module (FBAM)}

Our second attention module is simply obtained by transposing the channel dimension with the spatial dimensions before applying the same transformations.
In other words, we transpose the input $X'$ into $\mathcal{R}^{d_N \times d_C}$ and transpose the output $Y'$ back into $\mathcal{R}^{d_C \times d_N}$. While SAM models an attention over the $d_H*d_W$ regions that compose the input features map, FBAM models an attention over the $d_C$ features channels.

We obtain the best results with a representation space of 196 dimensions ($d=196$) and one attention head ($n_H=1$).

\begin{figure*}[t]
\centering
  \includegraphics[width=1\linewidth]{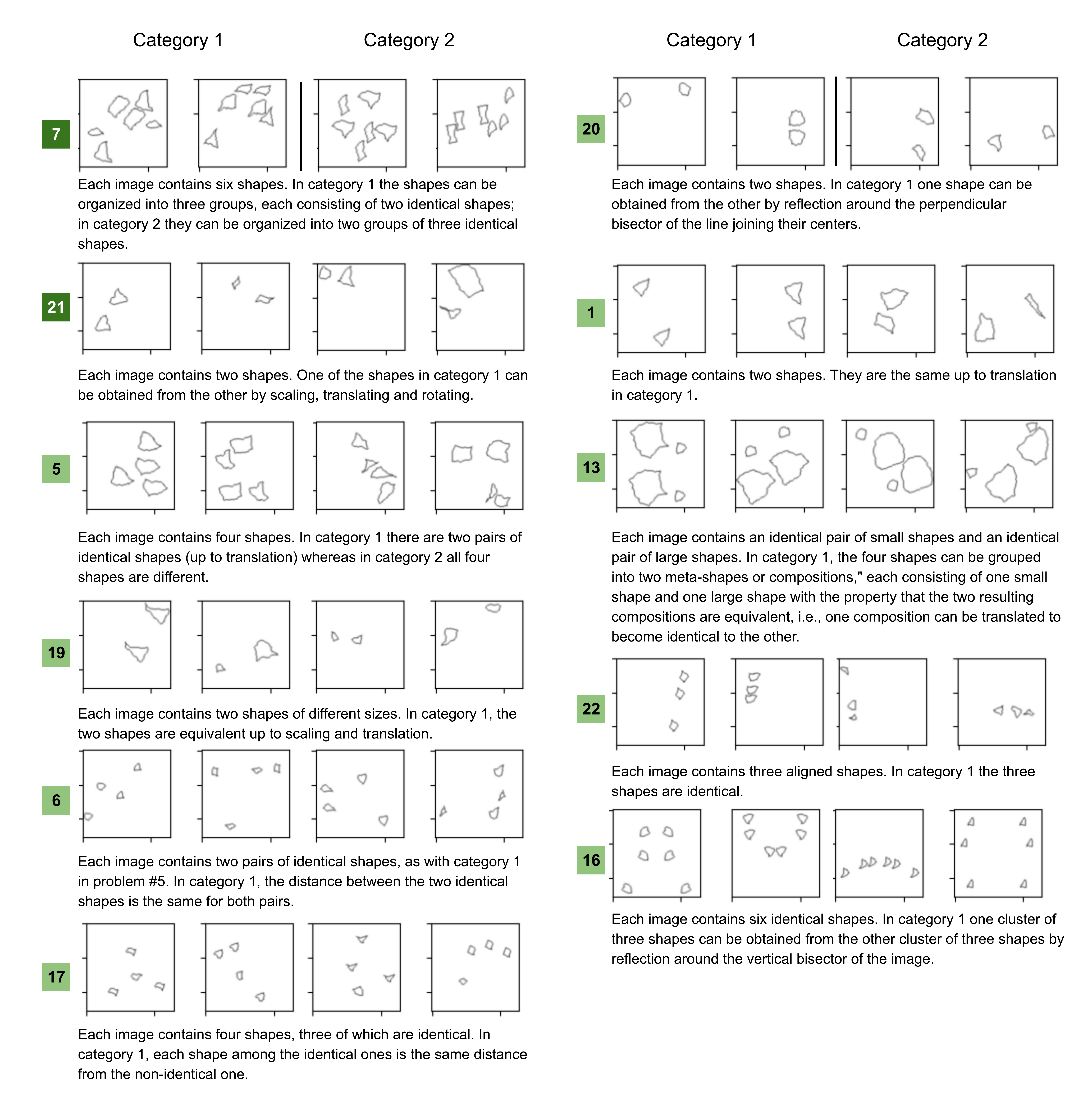}
 \caption{Sample images for Same Different (SD) tasks }\label{fig:exampleSD}
\end{figure*}

\begin{figure*}[t]
\centering
  \includegraphics[width=1\linewidth]{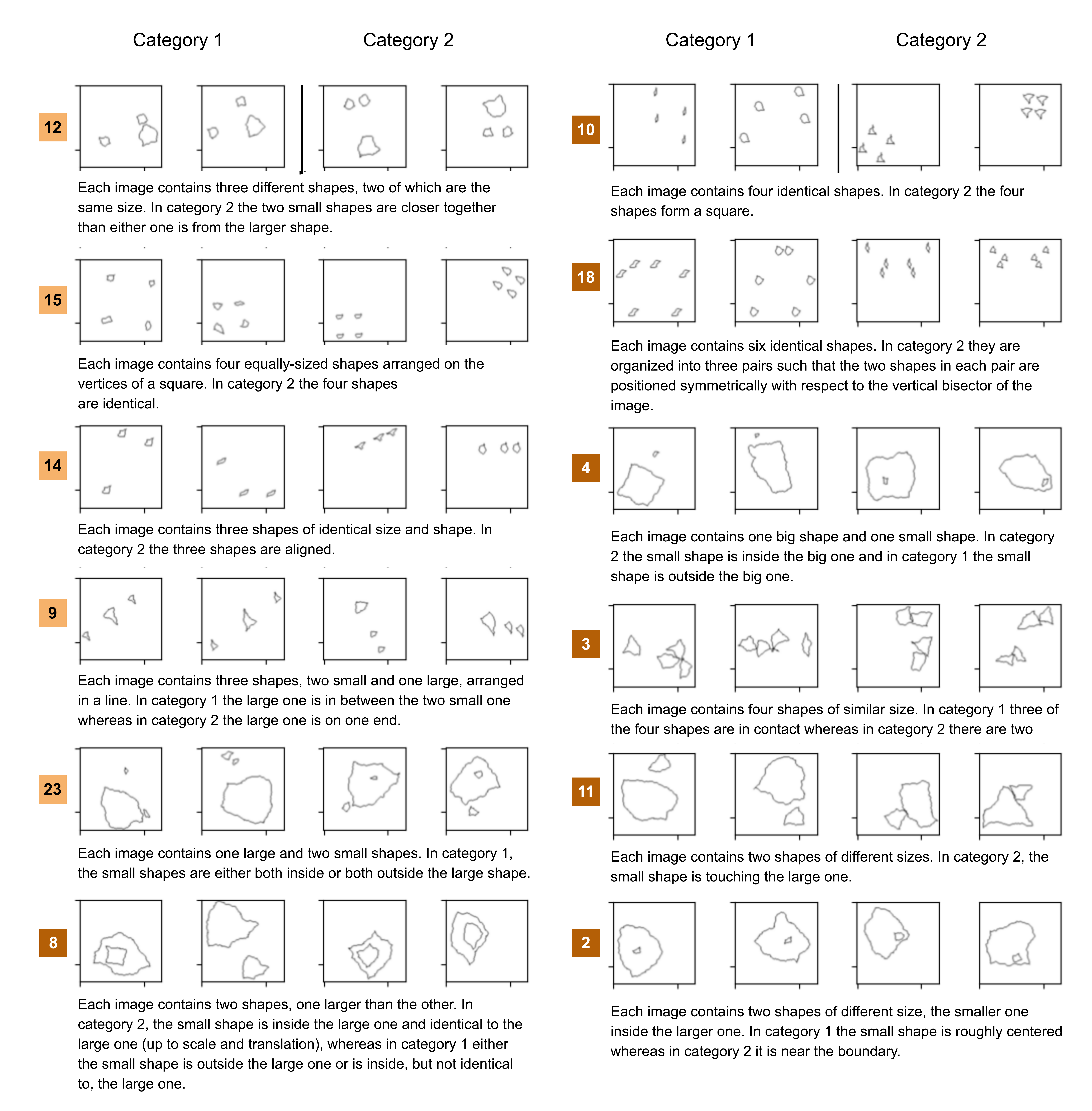}
 \caption{Sample images for Spatial Relation (SR) tasks }\label{fig:exampleSR}
\end{figure*}

\begin{table}[t]
\centering
\caption{Each cell represents attempts participants took to solve seven consecutive correct categorization. Here, row and column represents $task~ number$ and $participant ~ number$. Entries containing "X" indicate that the participant failed to solve the problem, and those cells are not included in the marginal means. \citep{fleuret2011comparing}}
\vspace{5mm}

\small
\begin{adjustbox}{width=\textwidth}
\begin{tabular}{|c|llllllllllllllllllll|c|c|}
\hline 
\multicolumn{1}{|l|}{} & \multicolumn{20}{c}{\textbf{Participant No.}}                                                                                                                                                                                                                                                                                                                                                                                                                                                                                                                                                                                                                                                                                                                                                                                                                                             & \multicolumn{1}{|l|}{}                                & \multicolumn{1}{|l|}{}                                \\ \cline{2-21}
\textbf{Task No.}    & \textbf{1}             & \textbf{2}                                 & \textbf{3}                                 & \textbf{4}                                 & \textbf{5}                                 & \textbf{6}                                 & \textbf{7}             & \textbf{8}                                 & \textbf{9}                                 & \textbf{10}                                & \textbf{11}                                & \textbf{12}                                & \textbf{13}                                & \textbf{14}                                & \textbf{15}                                & \textbf{16}                                & \textbf{17}                                & \textbf{18}                                & \textbf{19}                                & \textbf{20}                                & \multicolumn{1}{|l|}{\multirow{-2}{*}{\textbf{Mean}}} & \multicolumn{1}{|l|}{\multirow{-2}{*}{\textbf{Fail}}} \\ \hline
\textbf{1}           & {\color{blue} 1}  & {\color{blue} 12}                     & {\color{blue} 1}                      & {\color{blue} 2}                      & {\color{blue} 8}                      & {\color{blue} 8}                      & {\color{blue} 1}  & {\color{blue} 1}                      & {\color{red} X} & {\color{blue} 1}                      & {\color{blue} 14}                     & {\color{blue} 1}                      & {\color{blue} 4}                      & {\color{blue} 1}                      & {\color{blue} 1}                      & {\color{blue} 1}                      & {\color{blue} 2}                      & {\color{blue} 1}                      & {\color{blue} 1}                      & {\color{blue} 1}                      & 3.26                                                & 1                                                   \\
\textbf{2}           & {\color{blue} 3}  & {\color{blue} 1}                      & {\color{blue} 2}                      & {\color{blue} 2}                      & {\color{blue} 10}                     & {\color{blue} 19}                     & {\color{blue} 4}  & {\color{blue} 4}                      & {\color{blue} 14}                     & {\color{blue} 3}                      & {\color{blue} 2}                      & {\color{blue} 3}                      & {\color{blue} 21}                     & {\color{blue} 1}                      & {\color{blue} 1}                      & {\color{blue} 5}                      & {\color{blue} 3}                      & {\color{blue} 2}                      & {\color{blue} 22}                     & {\color{blue} 9}                      & 6.55                                                & 0                                                   \\
\textbf{3}           & {\color{blue} 7}  & {\color{blue} 1}                      & {\color{blue} 3}                      & {\color{blue} 1}                      & {\color{blue} 4}                      & {\color{blue} 3}                      & {\color{blue} 1}  & {\color{blue} 1}                      & {\color{blue} 7}                      & {\color{blue} 1}                      & {\color{blue} 6}                      & {\color{blue} 1}                      & {\color{blue} 1}                      & {\color{blue} 1}                      & {\color{blue} 4}                      & {\color{blue} 1}                      & {\color{blue} 1}                      & {\color{blue} 1}                      & {\color{blue} 4}                      & {\color{blue} 2}                      & 2.55                                                & 0                                                   \\
\textbf{4}           & {\color{blue} 1}  & {\color{blue} 6}                      & {\color{blue} 7}                      & {\color{blue} 1}                      & {\color{blue} 1}                      & {\color{blue} 3}                      & {\color{blue} 1}  & {\color{blue} 1}                      & {\color{blue} 1}                      & {\color{blue} 1}                      & {\color{blue} 3}                      & {\color{blue} 1}                      & {\color{blue} 1}                      & {\color{blue} 2}                      & {\color{blue} 1}                      & {\color{blue} 1}                      & {\color{blue} 7}                      & {\color{blue} 5}                      & {\color{blue} 7}                      & {\color{blue} 1}                      & 2.6                                                 & 0                                                   \\
\textbf{5}           & {\color{blue} 7}  & {\color{red} X} & {\color{blue} 1}                      & {\color{blue} 21}                     & {\color{blue} 8}                      & {\color{blue} 3}                      & {\color{blue} 1}  & {\color{blue} 5}                      & {\color{red} X} & {\color{blue} 1}                      & {\color{red} X} & {\color{blue} 9}                      & {\color{blue} 13}                     & {\color{blue} 1}                      & {\color{blue} 6}                      & {\color{blue} 2}                      & {\color{red} X} & {\color{blue} 8}                      & {\color{blue} 1}                      & {\color{blue} 7}                      & 5.88                                                & 4                                                   \\
\textbf{6}           & {\color{red} X}  & {\color{blue} 20}                     & {\color{red} X} & {\color{red} X} & {\color{blue} 27}                     & {\color{blue} 25}                     & {\color{blue} 12} & {\color{blue} 26}                     & {\color{red} X} & {\color{red} X} & {\color{blue} 3}                      & {\color{red} X} & {\color{red} X} & {\color{red} X} & {\color{blue} 4}                      & {\color{blue} 16}                     & {\color{red} X} & {\color{red} X} & {\color{red} X} & {\color{red} X} & 16.63                                               & 12                                                  \\
\textbf{7}           & {\color{blue} 1}  & {\color{red} X} & {\color{blue} 1}                      & {\color{red} X} & {\color{blue} 13}                     & {\color{blue} 8}                      & {\color{blue} 4}  & {\color{blue} 14}                     & {\color{red} X} & {\color{blue} 3}                      & {\color{blue} 8}                      & {\color{blue} 12}                     & {\color{blue} 7}                      & {\color{red} X} & {\color{blue} 1}                      & {\color{blue} 6}                      & {\color{blue} 1}                      & {\color{blue} 1}                      & {\color{blue} 14}                     & {\color{blue} 9}                      & 6.44                                                & 4                                                   \\
\textbf{8}           & {\color{blue} 7}  & {\color{blue} 6}                      & {\color{blue} 1}                      & {\color{blue} 14}                     & {\color{blue} 4}                      & {\color{blue} 14}                     & {\color{blue} 1}  & {\color{blue} 5}                      & {\color{blue} 1}                      & {\color{blue} 4}                      & {\color{blue} 8}                      & {\color{blue} 1}                      & {\color{blue} 1}                      & {\color{blue} 1}                      & {\color{blue} 13}                     & {\color{blue} 5}                      & {\color{blue} 3}                      & {\color{blue} 7}                      & {\color{blue} 4}                      & {\color{blue} 1}                      & 5.05                                                & 0                                                   \\
\textbf{9}           & {\color{blue} 4}  & {\color{blue} 24}                     & {\color{blue} 1}                      & {\color{blue} 16}                     & {\color{blue} 3}                      & {\color{blue} 1}                      & {\color{blue} 1}  & {\color{blue} 13}                     & {\color{red} X} & {\color{red} X} & {\color{blue} 4}                      & {\color{blue} 6}                      & {\color{red} X} & {\color{blue} 2}                      & {\color{blue} 7}                      & {\color{blue} 1}                      & {\color{blue} 3}                      & {\color{blue} 1}                      & {\color{blue} 5}                      & {\color{blue} 1}                      & 5.47                                                & 3                                                   \\
\textbf{10}          & {\color{blue} 1}  & {\color{blue} 8}                      & {\color{blue} 2}                      & {\color{blue} 2}                      & {\color{blue} 4}                      & {\color{blue} 1}                      & {\color{blue} 3}  & {\color{blue} 5}                      & {\color{red} X} & {\color{blue} 4}                      & {\color{blue} 1}                      & {\color{blue} 2}                      & {\color{blue} 16}                     & {\color{blue} 4}                      & {\color{blue} 4}                      & {\color{blue} 2}                      & {\color{blue} 1}                      & {\color{blue} 1}                      & {\color{blue} 4}                      & {\color{blue} 3}                      & 3.58                                                & 1                                                   \\
\textbf{11}          & {\color{blue} 4}  & {\color{blue} 2}                      & {\color{blue} 3}                      & {\color{blue} 1}                      & {\color{blue} 3}                      & {\color{blue} 1}                      & {\color{blue} 4}  & {\color{blue} 8}                      & {\color{blue} 1}                      & {\color{blue} 2}                      & {\color{blue} 1}                      & {\color{blue} 1}                      & {\color{blue} 1}                      & {\color{blue} 1}                      & {\color{blue} 1}                      & {\color{blue} 5}                      & {\color{blue} 2}                      & {\color{blue} 1}                      & {\color{blue} 1}                      & {\color{blue} 1}                      & 2.2                                                 & 0                                                   \\
\textbf{12}          & {\color{blue} 1}  & {\color{blue} 2}                      & {\color{blue} 8}                      & {\color{blue} 1}                      & {\color{blue} 9}                      & {\color{blue} 4}                      & {\color{blue} 8}  & {\color{blue} 4}                      & {\color{blue} 1}                      & {\color{blue} 7}                      & {\color{blue} 25}                     & {\color{blue} 2}                      & {\color{blue} 5}                      & {\color{blue} 2}                      & {\color{red} X} & {\color{blue} 2}                      & {\color{blue} 5}                      & {\color{red} X} & {\color{blue} 4}                      & {\color{blue} 1}                      & 5.06                                                & 2                                                   \\
\textbf{13}          & {\color{blue} 1}  & {\color{blue} 20}                     & {\color{blue} 5}                      & {\color{blue} 14}                     & {\color{red} X} & {\color{blue} 3}                      & {\color{blue} 1}  & {\color{blue} 13}                     & {\color{blue} 7}                      & {\color{blue} 10}                     & {\color{blue} 1}                      & {\color{blue} 13}                     & {\color{blue} 9}                      & {\color{blue} 5}                      & {\color{red} X} & {\color{blue} 3}                      & {\color{blue} 3}                      & {\color{blue} 2}                      & {\color{red} X} & {\color{blue} 1}                      & 6.53                                                & 3                                                   \\
\textbf{14}          & {\color{blue} 4}  & {\color{blue} 4}                      & {\color{blue} 1}                      & {\color{blue} 1}                      & {\color{blue} 3}                      & {\color{blue} 10}                     & {\color{blue} 2}  & {\color{red} X} & {\color{blue} 12}                     & {\color{blue} 14}                     & {\color{blue} 1}                      & {\color{blue} 19}                     & {\color{blue} 1}                      & {\color{blue} 3}                      & {\color{blue} 1}                      & {\color{blue} 1}                      & {\color{blue} 4}                      & {\color{blue} 8}                      & {\color{blue} 1}                      & {\color{blue} 2}                      & 4.84                                                & 1                                                   \\
\textbf{15}          & {\color{blue} 1}  & {\color{red} X} & {\color{blue} 1}                      & {\color{blue} 2}                      & {\color{blue} 2}                      & {\color{blue} 1}                      & {\color{blue} 1}  & {\color{blue} 1}                      & {\color{red} X} & {\color{blue} 5}                      & {\color{blue} 1}                      & {\color{blue} 2}                      & {\color{blue} 4}                      & {\color{blue} 1}                      & {\color{blue} 1}                      & {\color{blue} 18}                     & {\color{blue} 10}                     & {\color{blue} 3}                      & {\color{blue} 2}                      & {\color{blue} 1}                      & 3.17                                                & 2                                                   \\
\textbf{16}          & {\color{blue} 12} & {\color{blue} 18}                     & {\color{blue} 7}                      & {\color{red} X} & {\color{red} X}                      & {\color{blue} 2}                      & {\color{blue} 2}  & {\color{blue} 14}                     & {\color{red} X} & {\color{red} X} & {\color{blue} 28}                     & {\color{blue} 9}                      & {\color{blue} 13}                     & {\color{red} X} & {\color{blue} 22}                     & {\color{blue} 10}                     & {\color{red} X} & {\color{red} X} & {\color{red} X} & {\color{red} X} & 12.45                                               & 9                                                   \\
\textbf{17}          & {\color{blue} 14} & {\color{red} X} & {\color{blue} 6}                      & {\color{blue} 5}                      & {\color{blue} 2}                      & {\color{red} X} & {\color{blue} 21} & {\color{red} X} & {\color{red} X} & {\color{blue} 22}                     & {\color{red} X} & {\color{blue} 14}                     & {\color{red} X} & {\color{red} X} & {\color{red} X} & {\color{red} X} & {\color{blue} 13}                     & {\color{blue} 8}                      & {\color{blue} 28}                     & {\color{blue} 1}                      & 12.18                                               & 9                                                   \\
\textbf{18}          & {\color{blue} 5}  & {\color{blue} 17}                     & {\color{blue} 2}                      & {\color{red} X} & {\color{blue} 27}                     & {\color{blue} 5}                      & {\color{blue} 5}  & {\color{blue} 1}                      & {\color{red} X} & {\color{blue} 2}                      & {\color{red} X} & {\color{blue} 7}                      & {\color{blue} 19}                     & {\color{blue} 4}                      & {\color{blue} 1}                      & {\color{blue} 1}                      & {\color{blue} 5}                      & {\color{blue} 1}                      & {\color{blue} 1}                      & {\color{blue} 2}                      & 6.18                                                & 3                                                   \\
\textbf{19}          & {\color{blue} 2}  & {\color{blue} 10}                     & {\color{blue} 1}                      & {\color{blue} 11}                     & {\color{blue} 1}                      & {\color{blue} 3}                      & {\color{blue} 5}  & {\color{blue} 11}                     & {\color{blue} 8}                      & {\color{blue} 2}                      & {\color{blue} 4}                      & {\color{blue} 2}                      & {\color{blue} 17}                     & {\color{blue} 1}                      & {\color{blue} 4}                      & {\color{blue} 4}                      & {\color{blue} 1}                      & {\color{blue} 6}                      & {\color{blue} 1}                      & {\color{red} X} & 4.95                                                & 1                                                   \\
\textbf{20}          & {\color{blue} 14} & {\color{blue} 7}                      & {\color{blue} 4}                      & {\color{blue} 5}                      & {\color{blue} 1}                      & {\color{blue} 8}                      & {\color{blue} 3}  & {\color{blue} 1}                      & {\color{red} X} & {\color{blue} 18}                     & {\color{blue} 9}                      & {\color{blue} 16}                     & {\color{blue} 3}                      & {\color{blue} 1}                      & {\color{blue} 6}                      & {\color{blue} 1}                      & {\color{blue} 2}                      & {\color{blue} 1}                      & {\color{blue} 15}                     & {\color{blue} 1}                      & 6.11                                                & 1                                                   \\
\textbf{21}          & {\color{blue} 6}  & {\color{red} X} & {\color{blue} 1}                      & {\color{red} X} & {\color{blue} 1}                      & {\color{red} X} & {\color{blue} 23} & {\color{red} X} & {\color{red} X} & {\color{blue} 21}                     & {\color{blue} 28}                     & {\color{blue} 7}                      & {\color{blue} 26}                     & {\color{blue} 7}                      & {\color{blue} 15}                     & {\color{blue} 2}                      & {\color{blue} 17}                     & {\color{red} X} & {\color{blue} 16}                     & {\color{red} X} & 13.08                                               & 7                                                   \\
\textbf{22}          & {\color{blue} 1}  & {\color{blue} 9}                      & {\color{blue} 14}                     & {\color{blue} 1}                      & {\color{blue} 1}                      & {\color{blue} 4}                      & {\color{blue} 1}  & {\color{blue} 5}                      & {\color{blue} 21}                     & {\color{blue} 2}                      & {\color{blue} 1}                      & {\color{blue} 2}                      & {\color{blue} 5}                      & {\color{blue} 1}                      & {\color{blue} 6}                      & {\color{blue} 1}                      & {\color{blue} 4}                      & {\color{blue} 1}                      & {\color{blue} 1}                      & {\color{blue} 6}                      & 4.35                                                & 0                                                   \\
\textbf{23}          & {\color{blue} 1}  & {\color{blue} 1}                      & {\color{blue} 7}                      & {\color{blue} 22}                     & {\color{blue} 1}                      & {\color{blue} 1}                      & {\color{blue} 2}  & {\color{blue} 1}                      & {\color{blue} 6}                      & {\color{blue} 21}                     & {\color{blue} 2}                      & {\color{blue} 5}                      & {\color{blue} 4}                      & {\color{blue} 6}                      & {\color{blue} 4}                      & {\color{blue} 3}                      & {\color{blue} 1}                      & {\color{blue} 1}                      & {\color{blue} 6}                      & {\color{blue} 8}                      & 5.15                                                & 0                                                   \\ \hline
\textbf{Mean}        & 4.45                   & 9.33                                       & 3.59                                       & 6.78                                       & 6.33                                       & 6.05                                       & 4.65                   & 6.7                                        & 7.18                                       & 7.2                                        & 7.5                                        & 6.14                                       & 8.55                                       & 2.37                                       & 5.15                                       & 4.14                                       & 4.4                                        & 3.11                                       &   6.9                                         &                    3.05                        & \multicolumn{1}{|l}{}                                & \multicolumn{1}{l|}{}                                \\ \hline
\textbf{No of Fails} & 1                      & 5                                          & 1                                          & 5                                          & 2                                          & 2                                          & 0                      & 3                                          & 12                                         & 3                                          & 3                                          & 1                                          & 3                                          & 4                                          & 3                                          & 1                                          & 3                                          & 4                                          &       3                                     &    4                                        & \multicolumn{1}{|l}{}                                & \multicolumn{1}{l|}{}    \\ \hline                        
\end{tabular}
\end{adjustbox}

\label{table:humans}
\end{table}

\begin{figure*}[t]
\centering
  \includegraphics[width=1\linewidth]{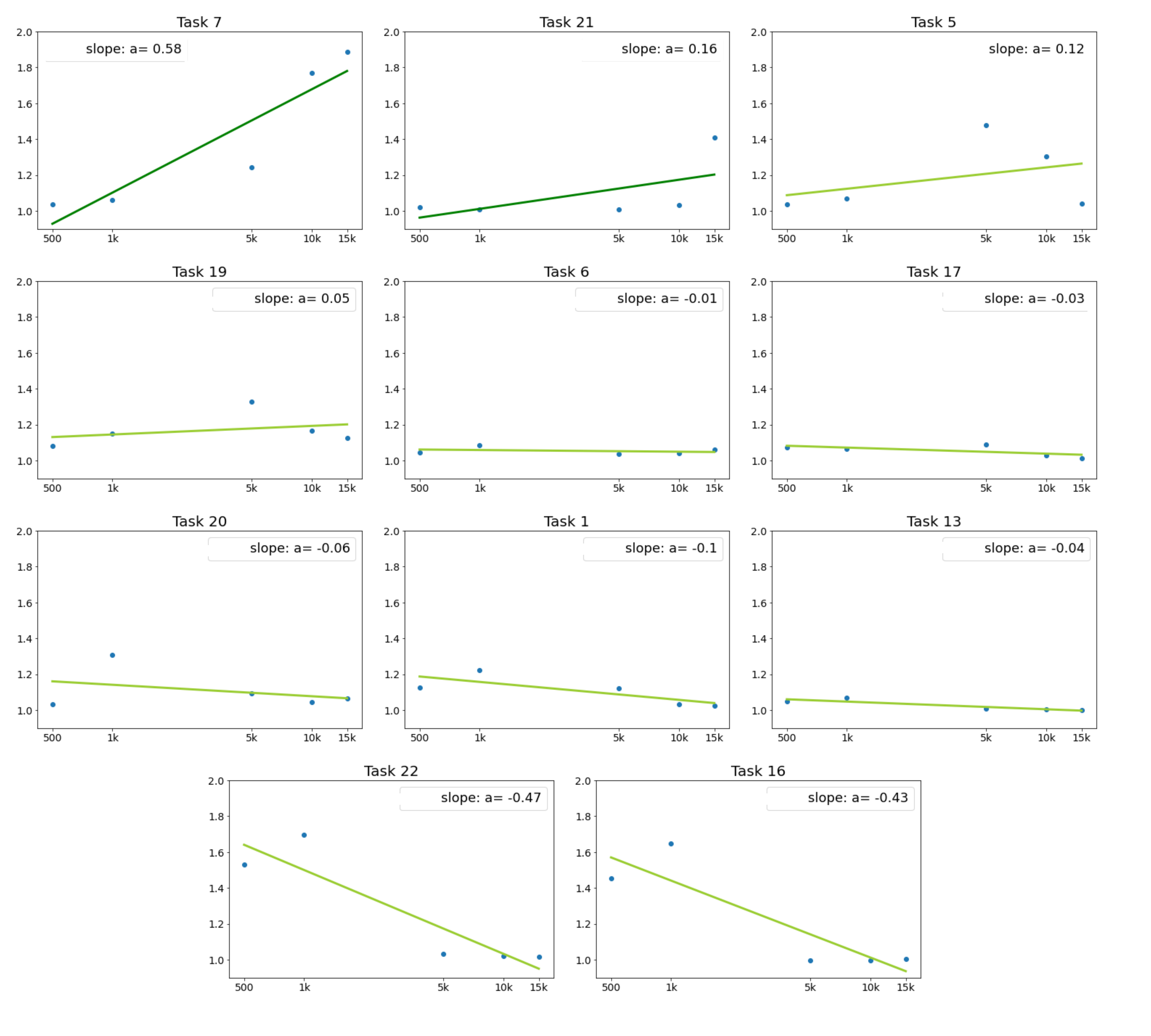}
 \caption{Slope attained by linear fitting of points obtained after taking the ratio of each of the network with spatial attention module and the test accuracy of a ResNet50 for each task and training condition for Same Different (SD) tasks }\label{fig:slope_sa_sd}
\end{figure*}

\begin{figure*}[t]
\centering
  \includegraphics[width=1\linewidth]{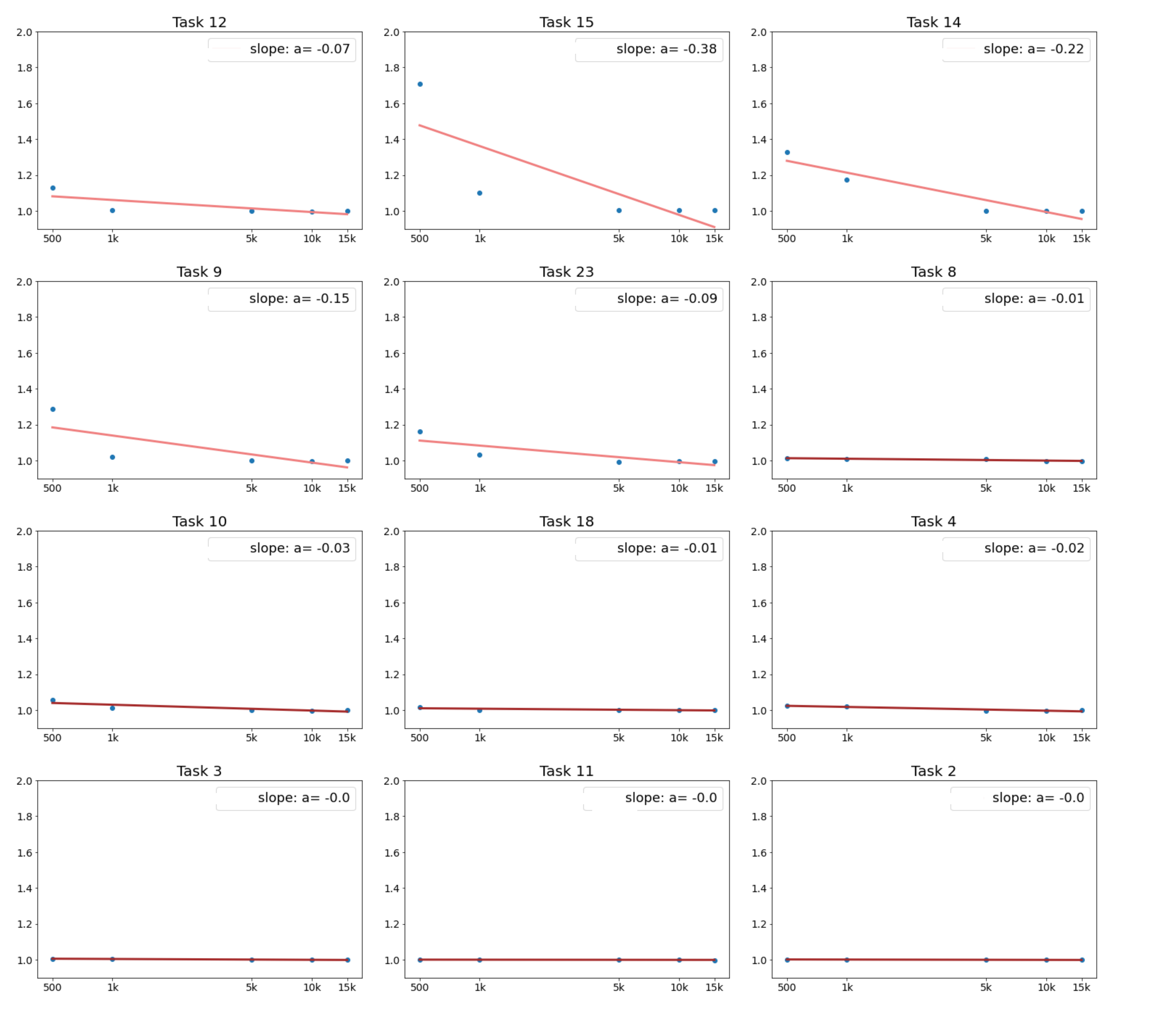}
 \caption{Slope attained by linear fitting of points obtained after taking the ratio of each of the network with spatial attention module and the test accuracy of a ResNet50 for each task and training condition for Spatial Relation (SR) tasks}\label{fig:slope_sa_sr}
\end{figure*}

\begin{figure*}[t]
\centering
  \includegraphics[width=1\linewidth]{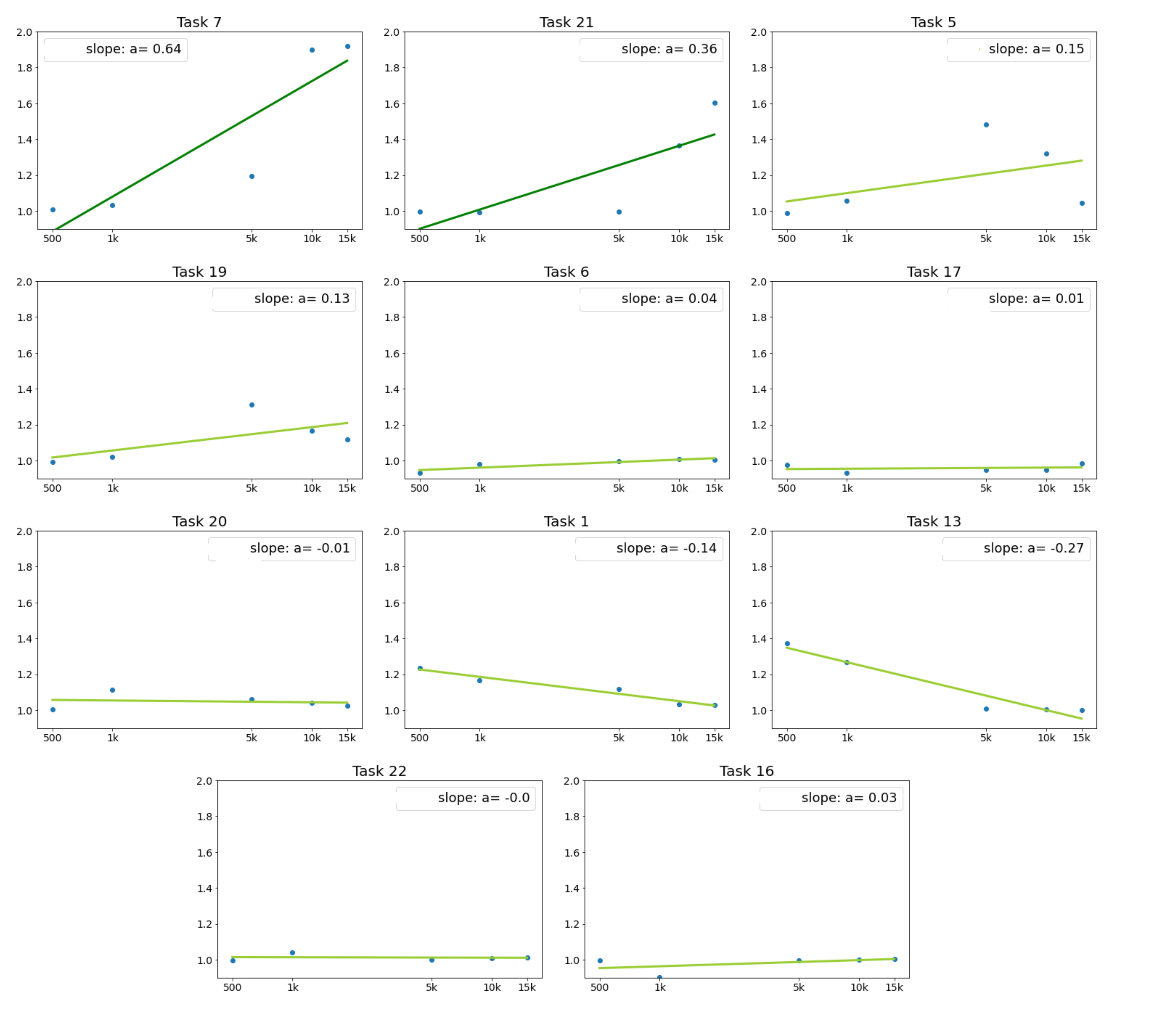}
 \caption{Slope attained by linear fitting of points obtained after taking the ratio of each of the network with feature-based attention module and the test accuracy of a ResNet50 for each task and training condition for Same Different (SD) tasks}\label{fig:slope_fba_sd}
\end{figure*}

\begin{figure*}[t]
\centering
  \includegraphics[width=1\linewidth]{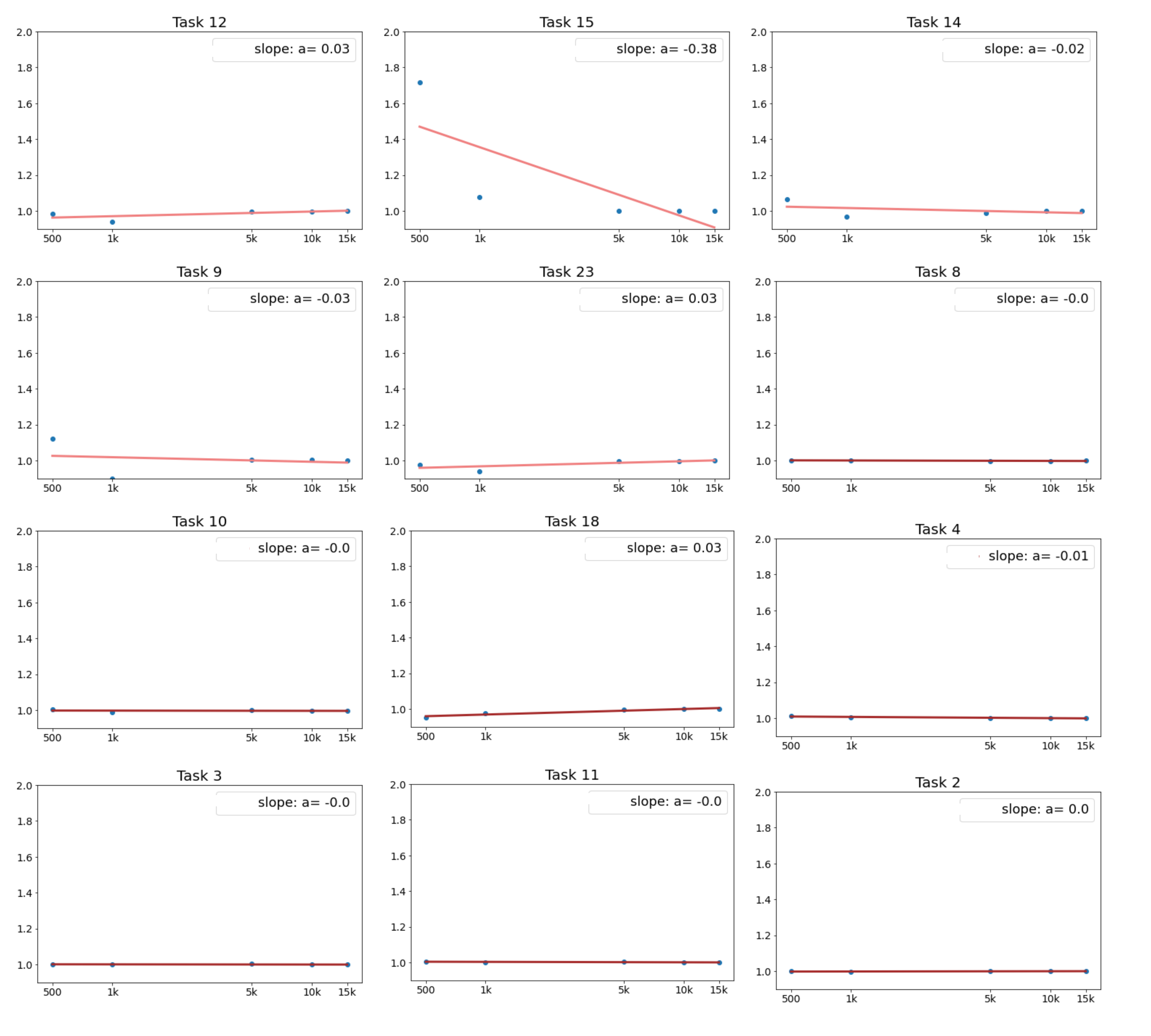}
 \caption{Slope attained by linear fitting of points obtained after taking the ratio of each of the network with feature-based attention module and the test accuracy of a ResNet50 for each task and training condition for Spatial Relation (SR) tasks }\label{fig:slope_fba_sr}
\end{figure*}

\begin{figure*}[htbp]
\centering
\includegraphics[width=1\linewidth]{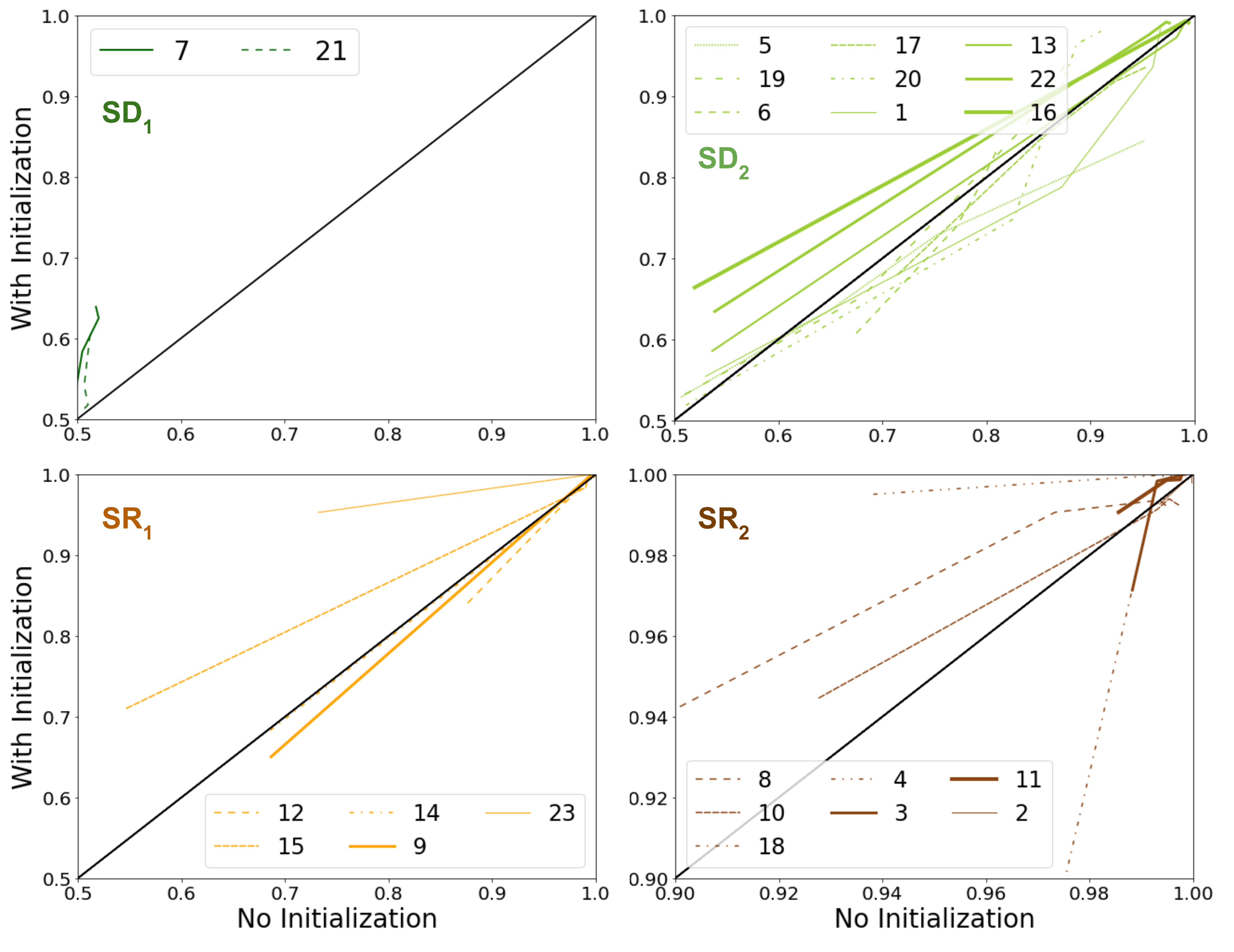} 
  
\caption{Test accuracies for a baseline ResNet50 trained from scratch (``No initialization'') vs. the same architecture pre-trained on Imagenet data for different number of training examples. The format is the same as used in Figure \ref{fig:xy_sa}. Also note that a different axis scale is used for $SR_2$ to improve visibility.}\label{fig:xy_img}
\end{figure*}

\end{document}